\documentclass{resona-report}
\makeatletter
\newcommand{\QueuePrimitiveDimensions}{%
  \begingroup
    \let\@xfloat\ReportKernelXFloat
    \let\ReportOriginalXFloat\@undefined
    \suppressfloats[t]%
    \begingroup
\ifdefined\ReportOriginalXFloat
  \par\leftskip=0pt\rightskip=0pt\noindent
  \begin{minipage}{\textwidth}
  \captionsetup{type=table}
\else
  \begin{table}[t]
\fi
\centering
\caption{\textbf{\textit{Auditory and visual primitive dimensions and the observable cues used to annotate the user input.}}}
\label{tab:primitives}
\ifdefined\ReportOriginalXFloat
  \begin{resonafigurecard}
\fi

\begingroup
\small
\setlength{\tabcolsep}{3pt}
\renewcommand{\arraystretch}{1.15}
\arrayrulecolor[gray]{0.45}
\setlength{\arrayrulewidth}{0.55pt}
\setlength{\cmidrulewidth}{0.45pt}

\begin{tabularx}{\linewidth}{%
  @{}l|>{\raggedright\arraybackslash}X@{}
}
\toprule[0.7pt]

\textbf{\textit{ID}}
& \textbf{\textit{Dimension and source-side evidence}}\\

\midrule[0.4pt]

\rowcolor{ClosedBand}
\multicolumn{2}{c}{%
  \textcolor{ClosedText}{%
    \textbf{\textit{Auditory primitives}}%
  }%
}\\
\midrule[0.25pt]

A1 & \textbf{Prosody:}
pitch, loudness, rate, stress, rhythm, pauses\\

A2 & \textbf{Speaker emotion:}
auditory evidence for affect or stance\\

A3 & \textbf{Speech function:}
question, complaint, sharing, challenge\\

A4 & \textbf{Nonverbal vocalizations:}
laughter, crying, sighs\\

\midrule[0.4pt]

\rowcolor{OpenBand}
\multicolumn{2}{c}{%
  \textcolor{OpenText}{%
    \textbf{\textit{Visual primitives}}%
  }%
}\\
\midrule[0.25pt]

V1 & \textbf{Facial behavior:}
visible facial actions\\

V2 & \textbf{Gaze:}
direction, target, and changes\\

V3 & \textbf{Head movements:}
nods, shakes, turns\\

V4 & \textbf{Arm and hand gestures:}
pointing, movements, self-touch\\

\bottomrule[0.7pt]
\end{tabularx}
\endgroup
\ifdefined\ReportOriginalXFloat
  \end{resonafigurecard}
  \end{minipage}\par
  \addvspace{\textfloatsep}
\else
  \end{table}
\fi
\endgroup
  \endgroup}
\makeatother
\AddToHook{cmd/subsection/before}{%
  \ifnum\value{section}=3\relax
    \ifnum\value{subsection}=0\relax\QueuePrimitiveDimensions\fi
  \fi}

\definecolor{AvatarHead}{RGB}{239,241,244}
\definecolor{ClosedBand}{RGB}{239,246,253}
\definecolor{ClosedBest}{RGB}{218,235,250}
\definecolor{ClosedScore}{RGB}{199,224,246}
\definecolor{ClosedText}{RGB}{47,83,118}

\definecolor{OpenBand}{RGB}{240,248,242}
\definecolor{OpenBest}{RGB}{220,239,225}
\definecolor{OpenScore}{RGB}{201,229,209}
\definecolor{OpenText}{RGB}{48,99,65}

\definecolor{AvatarHead}{RGB}{235,239,244}
\definecolor{OutcomeHead}{RGB}{236,232,246}
\definecolor{OutcomeSub}{RGB}{247,245,251}
\definecolor{ProcessHead}{RGB}{247,237,221}
\definecolor{ProcessSub}{RGB}{252,248,240}
\definecolor{OutcomeStar}{RGB}{133,105,180}
\definecolor{ProcessStar}{RGB}{194,145,40}
\definecolor{ScoreHead}{RGB}{239,241,244}

\DeclareRobustCommand{\closedkey}{%
  \colorbox{ClosedBest}{\rule{0pt}{0.8ex}\rule{0.8ex}{0pt}}%
}
\DeclareRobustCommand{\openkey}{%
  \colorbox{OpenBest}{\rule{0pt}{0.8ex}\rule{0.8ex}{0pt}}%
}
\definecolor{draftcolor}{RGB}{145,65,30}
\definecolor{outcomecolor}{RGB}{225,237,247}
\definecolor{evidencecolor}{RGB}{229,241,232}

\newcommand{\bench}{UUBench}
\usepackage{graphicx}
\usepackage{subcaption}
\colorlet{ClosedBand}{ResonaBlueSoft!70!white}
\colorlet{ClosedBest}{ResonaBlueSoft}
\colorlet{ClosedScore}{ResonaBlueMid}
\colorlet{ClosedText}{ResonaBlueDeep}
\colorlet{OpenBand}{ResonaOchreSoft!70!white}
\colorlet{OpenBest}{ResonaOchreSoft}
\colorlet{OpenScore}{ResonaOchreMid}
\colorlet{OpenText}{ResonaOchreDeep}
\colorlet{AvatarHead}{ResonaPanel}
\colorlet{OutcomeHead}{ResonaBlueSoft}
\colorlet{OutcomeSub}{ResonaBlueSoft!45!white}
\colorlet{ProcessHead}{ResonaOchreSoft}
\colorlet{ProcessSub}{ResonaOchreSoft!45!white}
\colorlet{OutcomeStar}{ResonaBlue}
\colorlet{ProcessStar}{ResonaOchre}
\colorlet{ScoreHead}{ResonaPanel}
\definecolor{draftcolor}{HTML}{A3453B}
\colorlet{outcomecolor}{ResonaBlueSoft}
\colorlet{evidencecolor}{ResonaOchreSoft}

\definecolor{UUBenchGold}{HTML}{C99415}

\reportbrand{RESONA}
\reportname{UUBench}
\title{%
  \textcolor{ResonaBlue}{U}\textcolor{UUBenchGold}{U}Bench: \textcolor{ResonaBlue}{U}nderstanding the \textcolor{UUBenchGold}{U}nspoken toward Full-Duplex Multimodal Interaction%
  \\[0.6em]
  {\normalfont\sffamily\fontsize{12}{16}\selectfont
   v0.1 Preview --- Work in Progress}%
}
\reportsubtitle{}
\reportversion{v0.1 Preview}
\reportstatus{TECHNICAL REPORT — WORK IN PROGRESS}
\reportwebsite{https://resona.ai/}
\reportdate{November 2026}
\abstract{
Effective real-time, full-duplex multimodal interaction requires systems to understand not only what users say, but also the communicative needs conveyed through prosody, facial expressions, gestures, and context, and to generate audiovisual responses that appropriately address those needs.
Existing evaluations offer limited insight into whether systems translate an understanding of such unspoken needs into effective responses.
We introduce \textbf{\textit{\bench{}}}, a benchmark for assessing how well multimodal interactive systems \textbf{\textit{U}}nderstand the \textbf{\textit{U}}nspoken intent and address users' implicit communicative needs.
The inherent subjectivity of implicit intent inference poses a key challenge for evaluation.
To address this, we introduce \textbf{\textit{Process-Oriented Scoring}}, which grounds the assessment of intent fulfillment in observable auditory and visual cues, such as prosody and gestures.
Rather than relying solely on a single overall score, it evaluates how well a response addresses the communicative needs conveyed by each applicable primitive dimension.
We aggregate the dimension-level scores according to each dimension's relevance to the annotated evidence.
Beyond intent fulfillment, \textbf{\textit{Outcome-Oriented Scoring}} evaluates the naturalness, appropriateness, and human-likeness of the reply itself under audio-only, visual-only, and audiovisual conditions.
To make these judgments traceable to observable evidence, we propose a \textbf{\textit{Fact-Grounded Scoring JEV}} module, which uses structured, timestamped audiovisual observations of the user and the digital human's reply to assess one dimension or criterion at a time.
UUBench provides a foundation for assessing whether multimodal interactive systems respond effectively to meaning beyond literal wording, supporting progress toward context-aware, real-time, full-duplex interaction.
}
\reportinstitutions{%
  \institutionrow{%
    \institutionlogo{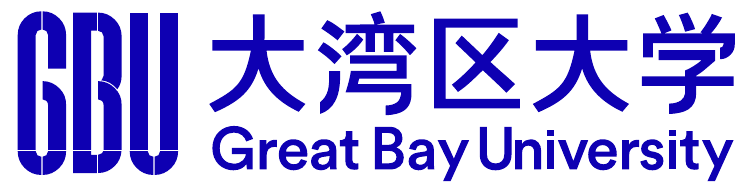}{https://www.gbu.edu.cn/}%
    \institutionlogo[width=0.9\linewidth,height=6.5mm]{assets/institutions/hkustgz.pdf}{https://www.hkust-gz.edu.cn/}%
    \institutionlogo[height=6.5mm]{assets/institutions/epfl.pdf}{https://www.epfl.ch/}%
    \institutionlogo[width=0.86\linewidth]{assets/institutions/cuhk.png}{https://www.cuhk.edu.hk/}%
  }%
  \institutionrow{%
        \institutionlogo[height=9mm]{assets/institutions/cumtb.png}{https://www.cumtb.edu.cn/}%
    \institutionlogo[trim=20.64bp 16.08bp 60.72bp 13.92bp,clip]{assets/institutions/kaust-full.png}{https://www.kaust.edu.sa/}%
    \institutionlogo{assets/institutions/kuleuven.pdf}{https://www.kuleuven.be/}%
    \institutionlogo{assets/institutions/kit.pdf}{https://www.kit.edu/}%
  }%
  \institutionrow{%
    \institutionlogo{assets/institutions/nagoya.pdf}{https://en.nagoya-u.ac.jp/}%
    \institutionlogo{assets/institutions/polito.pdf}{https://www.polito.it/en}%
    \institutionlogo[trim=46.32bp 40.32bp 43.92bp 46.32bp,clip]{assets/institutions/sjtu.png}{https://www.sjtu.edu.cn/}%
    \institutionlogo[trim=41bp 216bp 41bp 216bp,clip]{assets/institutions/astar.jpg}{https://www.a-star.edu.sg/}%
  }%
    \institutionrow{%
    \institutionlogo[width=0.9\linewidth,height=6.5mm]{assets/institutions/hkust.pdf}{https://www.hkust.edu.cn/}%
  }%
}

\newcommand{\authlead}[1]{{\sffamily\fontsize{11.5}{15}\selectfont\fontseries{sb}\selectfont\color{ResonaBlack}#1}}
\newcommand{\authname}[1]{\makebox[0.24\linewidth][l]{\color{ResonaInk}#1}\hskip 0pt\relax}
\newcommand{\authgap}{\hspace{6mm}}
\newcommand{\azpill}{\hspace{1.4mm}%
  \tikz[baseline=(t.base)]\node[fill=ResonaBlueSoft,rounded corners=1.4mm,
    inner xsep=1.4mm,inner ysep=0.8mm](t)
    {\sffamily\fontsize{5.8}{7}\selectfont\bfseries\color{ResonaBlueDeep}A--Z};}
\newcommand{\authhead}[1]{{\sffamily\fontsize{7.5}{10}\selectfont\color{ResonaMuted}#1}\par\vspace{1.2mm}}
\newcommand{\authblockgap}{\par\vspace{4.2mm}}
\newcommand{\ReportAuthorRows}{%
\frontrow{\eyebrow{Authors}}{\raggedright
    \vspace*{-1.5mm}%
    \authhead{Core contributors}
    \authlead{Ye Pan}\authgap\authlead{Chenfei Liao}%
    \authblockgap
    \authhead{Contributors\azpill}
    {\sffamily\fontsize{9}{14.5}\selectfont
    \authname{Dr. Bin Ren}%
    \authname{Hanqian Li}%
    \authname{Haocong He}%
    \authname{Dr. Jialei Chen}\par
    \authname{Jinlin He}%
    \authname{Dr. Kunyu Peng}%
    \authname{Mengyu Jin}%
    \authname{Dr. Shijie Li}\par
    \authname{Wenqian Cui}%
    \authname{Yuanqi Yao}%
    \authname{Dr. Yuqian Fu}\par}%
    \vspace{3mm}%
    \authhead{Academic Advisors}
    {\sffamily\fontsize{9}{14.5}\selectfont
    \authname{Dr. Irwin King}\par}%
    \vspace{3mm}%
\authhead{Industrial Collaborators}
\authlead{Kailong Ruan}\authgap\authlead{Yuan Gao}\authgap\authlead{Wei Wang}%
\authblockgap
\authhead{Correspondence}
\authlead{Dr. Xu Zheng}\authgap
{\sffamily\fontsize{8.5}{11}\selectfont
 \href{mailto:zhengx@gbu.edu.cn}{\color{ResonaBlue}zhengx@gbu.edu.cn}}\par}}
\makeatletter
\patchcmd{\mymaketitle}{\frontrow{\eyebrow{Abstract}}}{\@gobble}
  {}{\errmessage{Could not lift the abstract row}}
\patchcmd{\mymaketitle}{\makeinstitutionlogos}{%
  \ReportAuthorRows
  \makeinstitutionlogos
  \frontrow{\eyebrow{Abstract}}{\setlength{\parskip}{4pt}\fontsize{10.5}{14.6}\selectfont
    \leavevmode\abstractlist}%
}{}{\errmessage{Could not reorder the front rows}}
\makeatother

\renewcommand{\mymaketitle}{%
\hypersetup{pdftitle={UUBench: Understanding the Unspoken toward Full-Duplex Multimodal Interaction},pdfauthor={Ye Pan, Chenfei Liao, Bin Ren, Hanqian Li, Haocong He, Jialei Chen, Jinlin He, Kunyu Peng, Mengyu Jin, Shijie Li, Wenqian Cui, Yuanqi Yao, Yuqian Fu, Kailong Ruan, Yuan Gao, Wei Wang, Xuming Hu, Irwin King, Nicu Sebe, Xu Zheng},pdfcreator={XeLaTeX}}%
\begingroup\setlength{\leftskip}{0pt}\setlength{\rightskip}{0pt}
\begin{center}
{\sffamily\bfseries\fontsize{21}{25}\selectfont \textcolor{ResonaBlue}{U}\textcolor{UUBenchGold}{U}Bench:\\[1mm]\textcolor{ResonaBlue}{U}nderstanding the \textcolor{UUBenchGold}{U}nspoken toward\\Full-Duplex Multimodal Interaction\par}
\vspace{3mm}
\begin{minipage}{0.96\linewidth}\centering
{\sffamily\fontsize{10}{14}\selectfont
Ye Pan\textsuperscript{1,2,15,\ensuremath{\dagger}},
Chenfei Liao\textsuperscript{1,2,15,\ensuremath{\dagger}},
Bin Ren\textsuperscript{4},
Hanqian Li\textsuperscript{2},
Haocong He\textsuperscript{5},
Jialei Chen\textsuperscript{6},
Jinlin He\textsuperscript{7,15},
Kunyu Peng\textsuperscript{8},
Mengyu Jin\textsuperscript{9},
Shijie Li\textsuperscript{10},
Wenqian Cui\textsuperscript{11,15},
Yuanqi Yao\textsuperscript{12},
Yuqian Fu\textsuperscript{13},
\mbox{Kailong Ruan\textsuperscript{15,\ensuremath{\ddagger}}},
Yuan Gao\textsuperscript{15,\ensuremath{\ddagger}},
Wei Wang\textsuperscript{15,\ensuremath{\ddagger}},
Xuming Hu\textsuperscript{2,3,\ensuremath{\S}}, 
Irwin King\textsuperscript{11,\ensuremath{\S}}, Nicu Sebe\textsuperscript{14,\ensuremath{\S}},
Xu Zheng\textsuperscript{1,15,*}\par}
\vspace{1.5mm}
{\fontsize{8}{10.5}\selectfont
\mbox{\textsuperscript{1}Great Bay University}\enspace
\mbox{\textsuperscript{2}The Hong Kong University of Science and Technology (Guangzhou)}\\[1pt]
\mbox{\textsuperscript{3}The Hong Kong University of Science and Technology}\enspace
\mbox{\textsuperscript{4}EPFL}\enspace
\mbox{\textsuperscript{5}Shanghai Jiao Tong University}\enspace
\mbox{\textsuperscript{6}Nagoya University}\\[1pt]
\mbox{\textsuperscript{7}China University of Mining and Technology-Beijing}\enspace
\mbox{\textsuperscript{8}Karlsruhe Institute of Technology}\enspace
\mbox{\textsuperscript{9}Politecnico di Torino}\\[1pt]
\mbox{\textsuperscript{10}Agency for Science, Technology and Research (A*STAR), Singapore}\enspace
\mbox{\textsuperscript{11}The Chinese University of Hong Kong}\\[1pt]
\mbox{\textsuperscript{12}KU Leuven}\enspace
\mbox{\textsuperscript{13}King Abdullah University of Science and Technology}\enspace
\mbox{\textsuperscript{14}University of Trento}\enspace
\mbox{\textsuperscript{15}inSIght Labs}\par}
\vspace{1.5mm}
{\fontsize{8}{10}\selectfont
\textsuperscript{\ensuremath{\dagger}}Core contributors. Other contributors follow the report's A--Z order. \\
\textsuperscript{\ensuremath{\S}}Academic Advisors. 
\textsuperscript{\ensuremath{\ddagger}}Industrial collaborators.\par
\textsuperscript{*}Correspondence: \href{mailto:zhengx@gbu.edu.cn}{zhengx@gbu.edu.cn}\par}
\end{minipage}
\end{center}
\vspace{1mm}
{\fontsize{8}{10}\selectfont\color{ResonaMuted}\ReportVersion\enspace---\enspace Work in Progress\quad\ReportDate\par}
\vspace{2mm}
\noindent{\color{ResonaLine}\rule{\linewidth}{0.4pt}}\par\vspace{2mm}
\noindent{\sffamily\bfseries Abstract}\par\smallskip
{\fontsize{10.5}{14}\selectfont\abstractlist\par}
\vspace{2mm}\endgroup}

\begin{document}
\maketitle
\section{Introduction}
\label{sec:intro}
In everyday interactions, users often convey unstated communicative needs through language, intonation, facial expressions, and gestures. 
Listeners interpret these cues in context to infer an unstated communicative goal \citep{clark1979,kelly1999,hellbernd2016}. 
Thus, we define \textit{\textbf{Implicit Intent}} as a communicative goal that is not directly and fully stated in literal verbal content, but can be inferred from verbal cues, observable nonverbal behavior, and available context. 
For interactive digital humans, understanding such intent must be reflected in generated language, vocal expression, and visible behavior. 
Whether a digital human recognizes users' unstated needs and addresses them appropriately is therefore an important criterion for evaluating its interactive quality. 
In Case~I of Figure~\ref{fig:teaser}, Tavus focuses on whether ``I'm dead'' is a joke and misses the speaker's distress about quitting his job. In Case~II, Vidu responds playfully to another speaker's account of performing for her parents' guests, overlooking the sense of obligation in her words ``I had to do a play.'' Appendix~\ref{app:worked-evaluation} (\hyperref[app:worked-clip-95]{Case~D} and \hyperref[app:worked-clip-180]{Case~E}) shows how five digital human systems responded to these two user clips.

\begingroup
\makeatletter
\ifdefined\ReportOriginalXFloat
  \let\@xfloat\ReportOriginalXFloat
\fi
\makeatother
\begin{figure*}[!t]
\centering
\ifdefined\ReportOriginalIncludegraphics
  \ReportOriginalIncludegraphics[width=\textwidth]{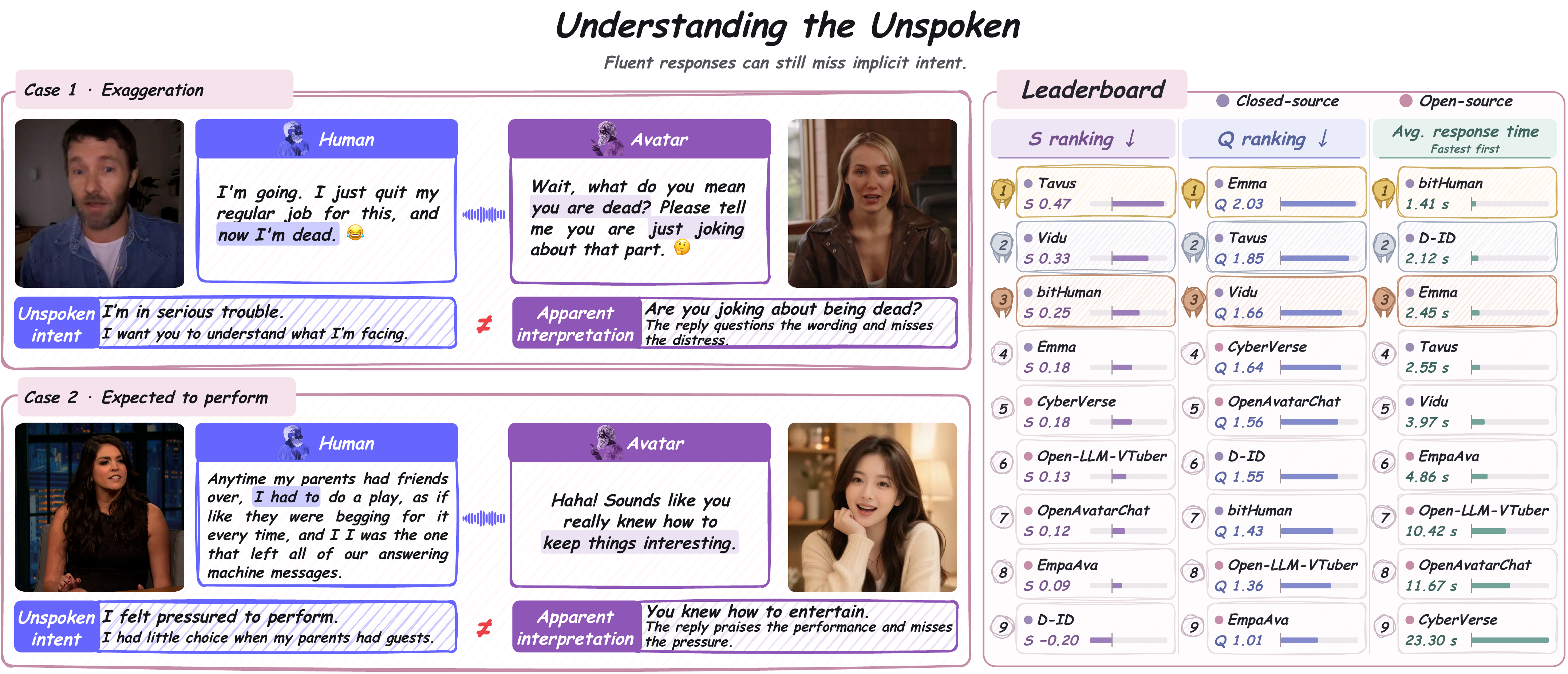}
\else
  \includegraphics[width=\textwidth]{figures/teaser-intent-gap.pdf}
\fi
\caption{\textbf{\textit{Two examples of missed intent.}} In Case~I, the reply asks whether ``I'm dead'' is literal when the speaker is describing a predicament. In Case~II, Vidu praises the speaker for entertaining her parents' guests but overlooks the pressure she describes. The leaderboard ranks nine systems by evidence-weighted intent fulfillment $S$ and outcome quality $Q$ on 216 shared inputs. Times are averaged per system. For closed-source systems, timing runs from the end of user-video playback until the avatar begins speaking. For open-source systems, timing covers offline reply generation and rendering.}
\label{fig:teaser}
\end{figure*}
\endgroup

\ifdefined\ReportOriginalXFloat
  \begingroup
\makeatletter
\ifdefined\ReportOriginalXFloat
  \let\@xfloat\ReportOriginalXFloat
\fi
\makeatother
\begin{figure*}[!t]
\centering
\includegraphics[width=\textwidth]{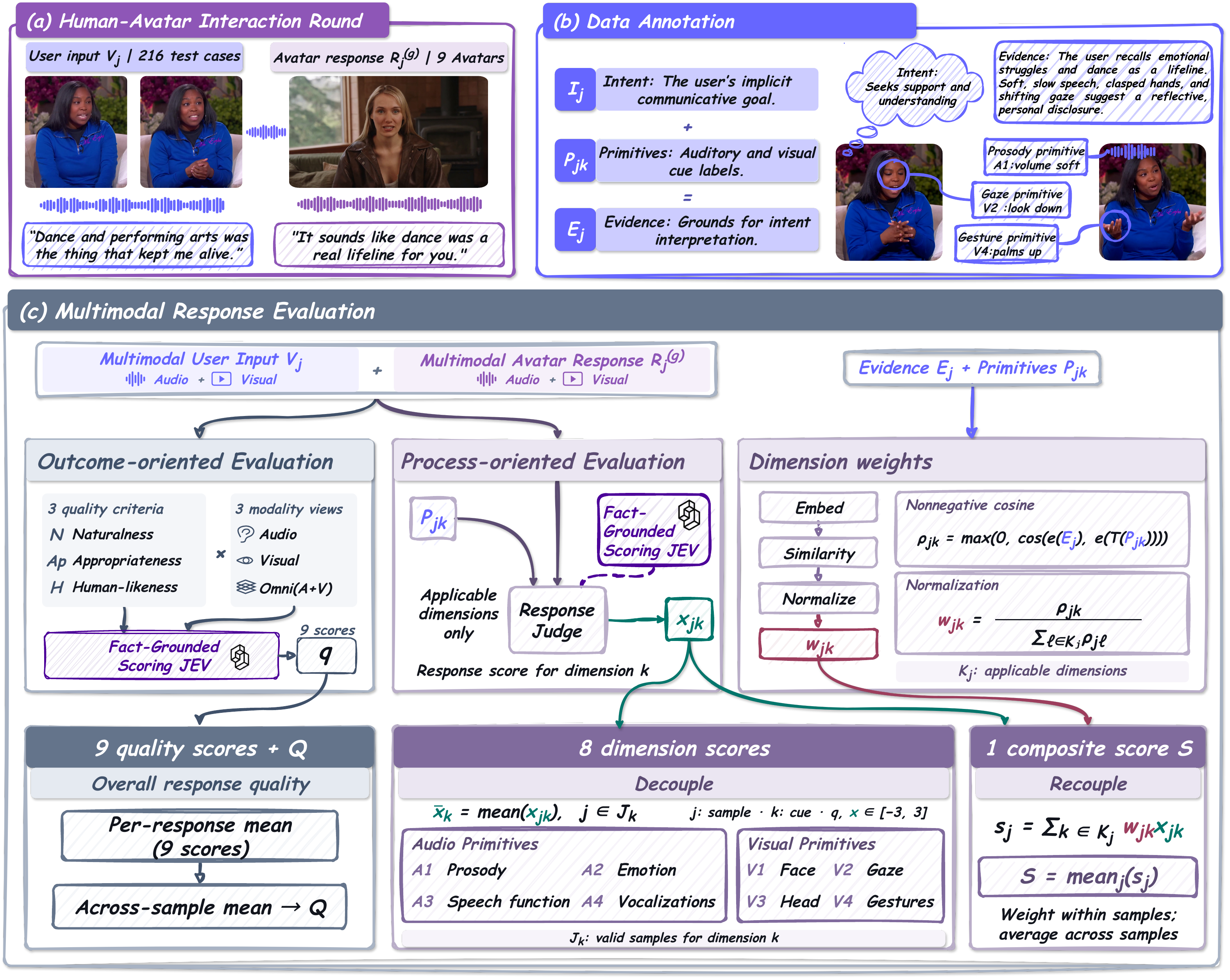}
\caption{\textbf{\textit{Evaluation pipeline of \bench{}.}} (a) A user describes dance as a lifeline, and an avatar responds. (b) The speaker's words, soft voice, lowered gaze, and upturned palms provide evidence that she seeks support and understanding. (c) Fact-Grounded Scoring JEV assesses audiovisual response quality ($Q$) and how well the response addresses the needs conveyed by each applicable primitive. The latter scores are reported by dimension ($\bar{x}_k$) and combined within each response using evidence-based weights to obtain the system score $S$ after averaging across clips.}
\label{fig:method}
\label{fig:framework}
\end{figure*}
\endgroup

\fi

Existing research provides evaluation foundations for both intent understanding and response generation. 
MIntRec2.0 assesses intent recognition through multimodal classification \citep{mintrec2024}, while VAGUE uses images to disambiguate what a speaker is asking someone to do \citep{vague2025}. 
For a subset of its clips, GesBench asks annotators what the speaker intends, how the listener might respond, and what led them to those judgments \citep{gesbench2026}. 
AvaMERG evaluates empathy in textual replies and the consistency of content and style across text, speech, and avatar video \citep{avamerg2025}. VidChatBench assesses dialogue coherence and audiovisual generation quality \citep{speakervid2025}, while LiveTalk evaluates response content, nonverbal engagement, and consistency across turns \citep{livetalk2025}.
However, these benchmarks do not systematically evaluate whether generated digital human responses fulfill the implicit communicative intent conveyed by users' multimodal cues.
A reply can sound natural and stay on topic while overlooking the speaker's need for acknowledgment. 
Evaluating such intent fulfillment requires explicitly linking the evidence supporting the inferred intent to the digital human's response.


To address this gap, we introduce \bench{}, a benchmark for evaluating how well digital humans respond to users' implicit communicative intentions.
Implicit intent is not directly observable, and inferring it from an interaction involves subjective judgment. 
A single rating of intent fulfillment can obscure which aspects of the user's need a reply addresses. 
We therefore introduce \textbf{\textit{Process-Oriented Scoring}}, which grounds this assessment in observable auditory and visual cues. 
As shown in Figure~\ref{fig:method}, the speaker describes how dance sustained her. 
The annotated evidence links her words, soft voice, and lowered gaze to an inferred need for support and understanding. 
We score the reply separately for each applicable primitive dimension based on how well it addresses the need conveyed by the corresponding cues. 
We then combine these scores using weights derived from the relevance of each primitive's labels to the annotated evidence. 
Beyond intent fulfillment, \textbf{\textit{Outcome-Oriented Scoring}} assesses the quality of the reply itself by rating its naturalness, appropriateness, and human-likeness under audio, visual, and joint audiovisual observation. 
To ground both types of scores in observable evidence, our \textbf{\textit{Fact-Grounded Scoring JEV}} module uses structured, timestamped audiovisual observations to evaluate one primitive dimension or quality criterion at a time.

We selected 28,548 candidate user clips from four video sources---\hfil\break TalkVerse~\citep{talkverse2025}, TalkCuts~\citep{talkcuts2025}, TalkVid~\citep{talkvid2025}, and SpeakerVid-5M~\citep{speakervid2025}.
Human reviewers then selected 216 high-quality clips from this pool for evaluation. 
For each selected clip, annotators labeled the speaker's intent and the auditory and visual primitives, then recorded the words and observable behaviors that supported their interpretation. 
We evaluated five closed-source and four open-source digital human systems on these same 216 inputs. 
All nine systems received positive overall audiovisual quality scores ($Q=1.01$--$2.03$), yet each had a negative mean score on every visual primitive dimension (Table~\ref{tab:overall}). 
These results suggest a shared difficulty among the evaluated systems in addressing communicative needs conveyed through facial behavior, gaze, head movements, and gestures.

Our main contributions are:
\begin{enumerate}
\setlength{\itemsep}{2pt}
\item \textbf{A benchmark for implicit intent fulfillment.} We introduce \bench{} to evaluate whether digital human responses address users' implicit communicative intent.
From a pool of 28,548 user clips, we selected 216 for evaluation. 
For each selected clip, annotators identified the speaker's intent, labeled the auditory and visual primitives, and recorded the words and behaviors supporting their interpretation.
\item \textbf{Scoring response quality and intent fulfillment.} 
We introduce two complementary scoring perspectives (Figure~\ref{fig:method}): 
\textbf{\textit{Process-Oriented Scoring}} assesses how well the reply addresses the need conveyed by each applicable primitive. 
\textbf{\textit{Outcome-Oriented Scoring}} measures audiovisual response quality ($Q$). 
Our \textbf{\textit{Fact-Grounded Scoring JEV}} module grounds both assessments in structured audio-visual observations.
We report primitive-level scores and aggregate them using evidence-based weights within each reply, then average across clips to obtain the system-level intent-fulfillment score ($S$).
\item \textbf{Comparative evaluation of responses to implicit intent.} 
We evaluated five closed-source and four open-source digital human systems with UUBench. 
All nine achieve positive overall audiovisual quality scores, yet each receives negative mean intent-fulfillment scores across all four visual primitive dimensions: facial behavior, gaze, head movements, and gestures (Table~\ref{tab:overall}).
These findings show that positive overall audiovisual quality scores can coexist with systematic shortcomings in addressing visually conveyed intent, motivating further research into how digital humans can translate multimodal intent understanding into appropriate responses.
\end{enumerate}

\section{Related Work}
\label{sec:related}

\ifdefined\ReportOriginalXFloat
  \def\PlaceDatasetComparison{\begingroup
\makeatletter
\ifdefined\ReportOriginalXFloat
  \let\@xfloat\ReportOriginalXFloat
\fi
\makeatother
\begin{table*}[!t]
\centering
\caption[Datasets for multimodal intent understanding and interactive responses]{\textbf{\textit{Datasets for multimodal intent understanding and interactive responses.}} We compare datasets by input modality, evaluation target, and the annotations available for evaluating a response. Intent denotes the speaker's goal, and Evidence denotes the cues that support its interpretation. For \bench{}, Response Ref. denotes the auditory and visual primitives annotated in the user input.}
\label{tab:comparison}

\begingroup
\fontsize{8}{10}\selectfont
\setlength{\tabcolsep}{3pt}
\renewcommand{\arraystretch}{1.15}
\arrayrulecolor[gray]{0.45}
\setlength{\arrayrulewidth}{0.55pt}
\setlength{\cmidrulewidth}{0.45pt}

\newcommand{\cmpYes}{\ensuremath{\checkmark}}
\newcommand{\cmpNo}{\ensuremath{\times}}
\newcommand{\cmpRelated}{\ensuremath{\triangle}}

\begin{tabularx}{\textwidth}{%
  @{}l|c|>{\raggedright\arraybackslash}p{0.23\linewidth}|*{3}{>{\centering\arraybackslash}X}@{}
}
\toprule[0.7pt]

\multirow{2}{*}[-2pt]{\textbf{\textit{Dataset}}}
& \multirow{2}{*}[-2pt]{\textbf{\textit{Input}}}
& \multirow{2}{*}[-2pt]{\textbf{\textit{Evaluation Target}}}
& \multicolumn{3}{c}{%
    \textbf{\textit{Annotations}}}\\

\cmidrule(lr){4-6}

& & &
\textit{Intent}
& \textit{Evidence}
& \textit{Response Ref.}\\

\midrule[0.4pt]

\rowcolor{ClosedBand}
\multicolumn{6}{c}{%
  \textcolor{ClosedText}{%
    \textbf{\textit{Multimodal intent and pragmatic understanding}}%
  }%
}\\
\midrule[0.25pt]

MIntRec \citep{mintrec2022}
& T,A,V & Intent interpretation
& \cmpYes & \cmpNo & \cmpNo\\

MIntRec2.0 \citep{mintrec2024}
& T,A,V & Intent interpretation
& \cmpYes & \cmpNo & \cmpNo\\

VAGUE \citep{vague2025}
& T,I & Intent interpretation
& \cmpYes & \cmpNo & \cmpNo\\

READI \citep{readi2026}
& T,I & Intent interpretation
& \cmpYes & \cmpNo & \cmpNo\\

GesBench \citep{gesbench2026}
& A,V & Intent interpretation
& \cmpYes\textsuperscript{*}
& \cmpYes\textsuperscript{*}
& \cmpYes\textsuperscript{*}\\

\midrule[0.4pt]

\rowcolor{OpenBand}
\multicolumn{6}{c}{%
  \textcolor{OpenText}{%
    \textbf{\textit{Interactive response generation}}%
  }%
}\\
\midrule[0.25pt]

ViCo \citep{vico2022}
& A,V & Nonverbal reaction
& \cmpNo & \cmpNo & \cmpYes\\

MARS \citep{react2025}
& A,V & Nonverbal reaction
& \cmpNo & \cmpNo & \cmpYes\\

AvaMERG \citep{avamerg2025}
& T,A,V & Audiovisual response
& \cmpRelated & \cmpRelated & \cmpYes\\

VidChatBench \citep{speakervid2025}
& A,V & Audiovisual response
& \cmpNo & \cmpNo & \cmpYes\\

\midrule[0.4pt]

\rowcolor{OutcomeHead}
\multicolumn{6}{c}{%
  \textcolor{OutcomeStar}{%
    \textbf{\textit{Implicit-intent response evaluation}}%
  }%
}\\
\midrule[0.25pt]

\textbf{\bench{} (Ours)}
& \textbf{A,V}
& \textbf{Audiovisual response}
& \cmpYes & \cmpYes & \cmpYes\\

\bottomrule[0.7pt]
\end{tabularx}
\endgroup

\vspace{3pt}
\begin{minipage}{\textwidth}
\fontsize{7.5}{9}\selectfont
T/A/I/V: text/audio/image/video. A check mark indicates that the annotation is available.
\textsuperscript{*}GesBench provides these annotations for a subset of its clips. Its response references describe possible listener actions in text.
$\triangle$: related annotations---response goals and emotion causes
(AvaMERG).
$\times$: not reported under these definitions.
\end{minipage}
\end{table*}
\endgroup
}
\else
  \begingroup
\makeatletter
\ifdefined\ReportOriginalXFloat
  \let\@xfloat\ReportOriginalXFloat
\fi
\makeatother
\begin{figure*}[!t]
\centering
\includegraphics[width=\textwidth]{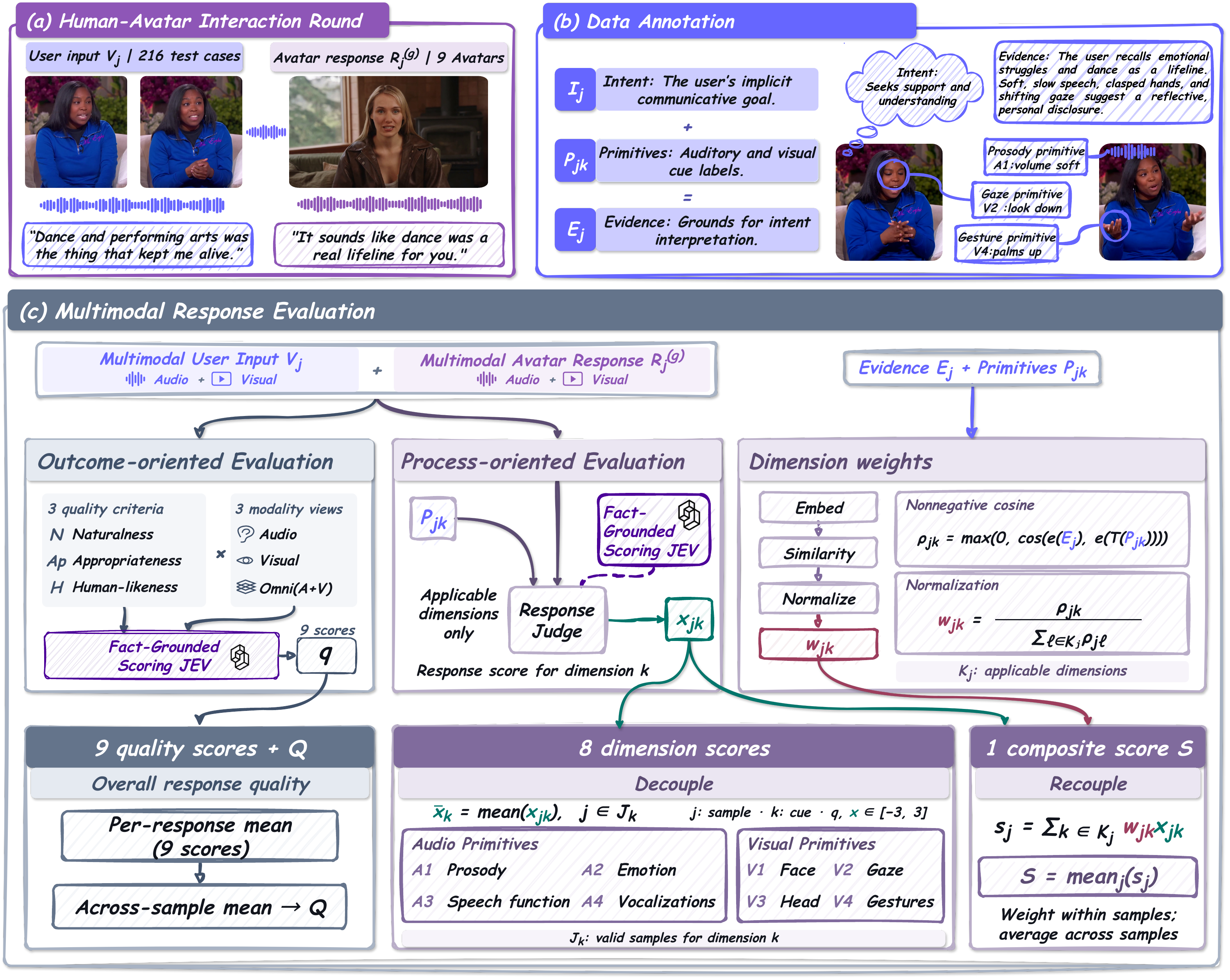}
\caption{\textbf{\textit{Evaluation pipeline of \bench{}.}} (a) A user describes dance as a lifeline, and an avatar responds. (b) The speaker's words, soft voice, lowered gaze, and upturned palms provide evidence that she seeks support and understanding. (c) Fact-Grounded Scoring JEV assesses audiovisual response quality ($Q$) and how well the response addresses the needs conveyed by each applicable primitive. The latter scores are reported by dimension ($\bar{x}_k$) and combined within each response using evidence-based weights to obtain the system score $S$ after averaging across clips.}
\label{fig:method}
\label{fig:framework}
\end{figure*}
\endgroup

\fi

\subsection{Multimodal Intent Recognition and Pragmatic Understanding}
Multimodal intent benchmarks examine how language, nonverbal behavior, and context inform communicative interpretation. MIntRec provides utterance-level intent classification \citep{mintrec2022}, while MIntRec2.0 extends this setting to multi-party conversations and out-of-scope detection \citep{mintrec2024}. VAGUE pairs an ambiguous expression with an image and asks what action the speaker is requesting \citep{vague2025}. READI tests whether models can use an image to infer a speaker's directive intent across three levels of indirectness in the utterance \citep{readi2026}. GesBench examines how interpretations change as an utterance unfolds. For 590 of its 6,208 clips, annotators describe the speaker's intent, suggest how a listener might respond, and explain their judgments \citep{gesbench2026}. Table~\ref{tab:comparison} summarizes the references these tasks provide. Predicting what a listener might do and evaluating how well an actual reply addresses the speaker's need are distinct tasks.

\ifdefined\ReportOriginalXFloat\else\newpage\fi
\ifdefined\begingroup
\makeatletter
\ifdefined\ReportOriginalXFloat
  \let\@xfloat\ReportOriginalXFloat
\fi
\makeatother
\begin{table*}[!t]
\centering
\caption[Datasets for multimodal intent understanding and interactive responses]{\textbf{\textit{Datasets for multimodal intent understanding and interactive responses.}} We compare datasets by input modality, evaluation target, and the annotations available for evaluating a response. Intent denotes the speaker's goal, and Evidence denotes the cues that support its interpretation. For \bench{}, Response Ref. denotes the auditory and visual primitives annotated in the user input.}
\label{tab:comparison}

\begingroup
\fontsize{8}{10}\selectfont
\setlength{\tabcolsep}{3pt}
\renewcommand{\arraystretch}{1.15}
\arrayrulecolor[gray]{0.45}
\setlength{\arrayrulewidth}{0.55pt}
\setlength{\cmidrulewidth}{0.45pt}

\newcommand{\cmpYes}{\ensuremath{\checkmark}}
\newcommand{\cmpNo}{\ensuremath{\times}}
\newcommand{\cmpRelated}{\ensuremath{\triangle}}

\begin{tabularx}{\textwidth}{%
  @{}l|c|>{\raggedright\arraybackslash}p{0.23\linewidth}|*{3}{>{\centering\arraybackslash}X}@{}
}
\toprule[0.7pt]

\multirow{2}{*}[-2pt]{\textbf{\textit{Dataset}}}
& \multirow{2}{*}[-2pt]{\textbf{\textit{Input}}}
& \multirow{2}{*}[-2pt]{\textbf{\textit{Evaluation Target}}}
& \multicolumn{3}{c}{%
    \textbf{\textit{Annotations}}}\\

\cmidrule(lr){4-6}

& & &
\textit{Intent}
& \textit{Evidence}
& \textit{Response Ref.}\\

\midrule[0.4pt]

\rowcolor{ClosedBand}
\multicolumn{6}{c}{%
  \textcolor{ClosedText}{%
    \textbf{\textit{Multimodal intent and pragmatic understanding}}%
  }%
}\\
\midrule[0.25pt]

MIntRec \citep{mintrec2022}
& T,A,V & Intent interpretation
& \cmpYes & \cmpNo & \cmpNo\\

MIntRec2.0 \citep{mintrec2024}
& T,A,V & Intent interpretation
& \cmpYes & \cmpNo & \cmpNo\\

VAGUE \citep{vague2025}
& T,I & Intent interpretation
& \cmpYes & \cmpNo & \cmpNo\\

READI \citep{readi2026}
& T,I & Intent interpretation
& \cmpYes & \cmpNo & \cmpNo\\

GesBench \citep{gesbench2026}
& A,V & Intent interpretation
& \cmpYes\textsuperscript{*}
& \cmpYes\textsuperscript{*}
& \cmpYes\textsuperscript{*}\\

\midrule[0.4pt]

\rowcolor{OpenBand}
\multicolumn{6}{c}{%
  \textcolor{OpenText}{%
    \textbf{\textit{Interactive response generation}}%
  }%
}\\
\midrule[0.25pt]

ViCo \citep{vico2022}
& A,V & Nonverbal reaction
& \cmpNo & \cmpNo & \cmpYes\\

MARS \citep{react2025}
& A,V & Nonverbal reaction
& \cmpNo & \cmpNo & \cmpYes\\

AvaMERG \citep{avamerg2025}
& T,A,V & Audiovisual response
& \cmpRelated & \cmpRelated & \cmpYes\\

VidChatBench \citep{speakervid2025}
& A,V & Audiovisual response
& \cmpNo & \cmpNo & \cmpYes\\

\midrule[0.4pt]

\rowcolor{OutcomeHead}
\multicolumn{6}{c}{%
  \textcolor{OutcomeStar}{%
    \textbf{\textit{Implicit-intent response evaluation}}%
  }%
}\\
\midrule[0.25pt]

\textbf{\bench{} (Ours)}
& \textbf{A,V}
& \textbf{Audiovisual response}
& \cmpYes & \cmpYes & \cmpYes\\

\bottomrule[0.7pt]
\end{tabularx}
\endgroup

\vspace{3pt}
\begin{minipage}{\textwidth}
\fontsize{7.5}{9}\selectfont
T/A/I/V: text/audio/image/video. A check mark indicates that the annotation is available.
\textsuperscript{*}GesBench provides these annotations for a subset of its clips. Its response references describe possible listener actions in text.
$\triangle$: related annotations---response goals and emotion causes
(AvaMERG).
$\times$: not reported under these definitions.
\end{minipage}
\end{table*}
\endgroup
\fi \subsection{Interactive Digital Human Generation}
Interactive digital human generation combines conversational behavior with audiovisual synthesis. INFP uses both participants' audio to animate an avatar's head and face as it speaks and listens \citep{infp2025}, while Avatar Forcing conditions causal generation on user audio and motion together with avatar speech to support real-time reactions \citep{avatarforcing2026}. Sustained interaction also requires temporal consistency and behavioral control. SoulX-LiveAct enables hour-scale, real-time animation with bounded historical memory and controllable expressions and actions \citep{liveact2026}. At the conversational-system level, LiveTalk connects an audio language model with streaming video generation for multi-turn interaction \citep{livetalk2025}, while InteractiveAvatar uses transcribed user speech and the avatar's current state to plan a spoken reply and accompanying actions \citep{interactiveavatar2026}. These developments make coordinated speech and visible feedback central to digital human interaction. Evaluating such systems therefore requires assessing both the quality of audiovisual realization and whether the generated behavior appropriately responds to the user's communicative goal.

\ifdefined\ReportOriginalXFloat\else\newpage\fi
\subsection{Evaluation of Digital Human Responses}
\label{sec:related-eval}\ifdefined\ReportOriginalXFloat\begingroup
\makeatletter
\ifdefined\ReportOriginalXFloat
    \let\@xfloat\ReportKernelXFloat
  \suppressfloats[t]
\fi
\makeatother
\begin{figure*}[t]
\centering
\includegraphics[width=\linewidth]{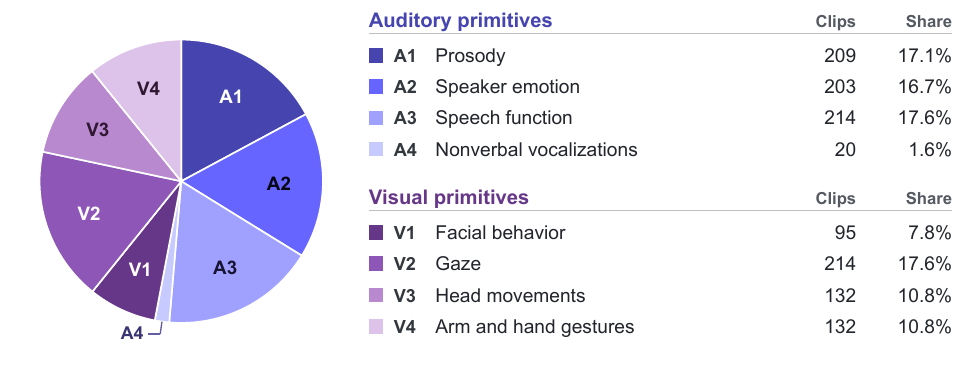}
\caption{\textbf{\textit{Primitive applicability across the 216 evaluation clips.}} Each clip contributes once to every dimension marked applicable. Counts give the number of clips for each dimension. Slice sizes and percentages show each dimension's share of all 1,219 clip--dimension assignments.}
\label{fig:curation}
\end{figure*}
\endgroup
\fi
Existing evaluations address several aspects of this problem. THEval measures talking-head quality, naturalness, and synchronization \citep{theval2025}. ViCo evaluates responsive listening behavior using paired speaker--listener recordings \citep{vico2022}; MARS, introduced for REACT 2025, supports multiple acceptable facial reactions and evaluates appropriateness, diversity, synchrony, and realism \citep{react2025}. AvaMERG annotates emotion causes and response goals to support empathetic reasoning. Its human evaluation rates empathy in textual replies and assesses the consistency of content and style across text, speech, and avatar video \citep{avamerg2025}. VidChatBench evaluates dialogue coherence alongside visual quality, identity preservation, audiovisual synchronization, and the alignment of facial expressions in generated and reference videos \citep{speakervid2025}. LiveTalk evaluates the semantic relevance of replies, nonverbal engagement, and visual and behavioral consistency across turns \citep{livetalk2025}. 

\bench{} links each annotated intent to observable cues in the user's clip. For each applicable auditory or visual primitive, it scores how well the generated reply addresses the need those cues convey. For example, a user describing a friend's move abroad as ``a great opportunity for her'' may speak at a lower pitch and volume, shift their gaze downward, turn down the corners of their mouth, and bow their head. A digital human should interpret these cues in context and recognize that the user may be seeking acknowledgment of their reluctance to part and disappointment about the separation.

\ifdefined\ReportOriginalXFloat\else
  \begingroup
\makeatletter
\ifdefined\ReportOriginalXFloat
    \let\@xfloat\ReportKernelXFloat
  \suppressfloats[t]
\fi
\makeatother
\begin{figure*}[t]
\centering
\includegraphics[width=\linewidth]{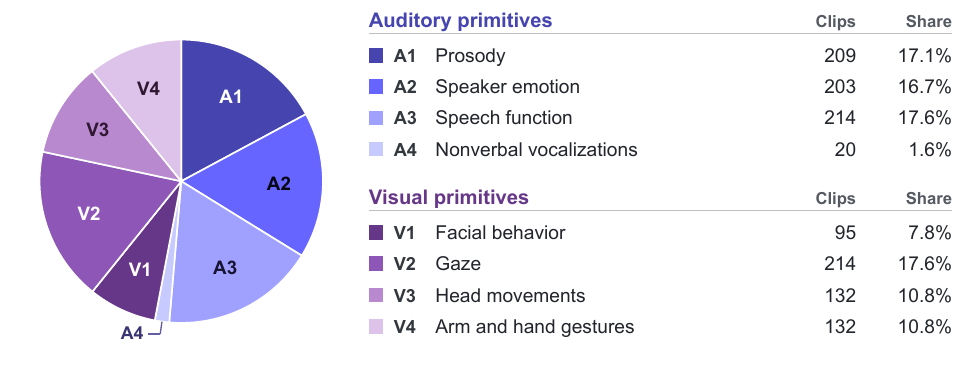}
\caption{\textbf{\textit{Primitive applicability across the 216 evaluation clips.}} Each clip contributes once to every dimension marked applicable. Counts give the number of clips for each dimension. Slice sizes and percentages show each dimension's share of all 1,219 clip--dimension assignments.}
\label{fig:curation}
\end{figure*}
\endgroup

  \newpage
\fi
\section{UUBench}
\label{sec:benchmark}

\subsection{Task Definition and Benchmark Overview}
\label{sec:task}
We define \emph{implicit communicative intent} as a speaker's communicative goal that they seek to make recognizable through expression in a particular interaction, but that is not directly and fully stated in the literal verbal content and must instead be inferred from relevant evidence in expression, observable nonverbal behavior, and available context \citep{clark1979,sperber2002,kelly1999,hellbernd2016}.

A visible cue such as lowered gaze can support an interpretation of the speaker's stance. Inferring a communicative goal, such as seeking reassurance, also requires considering what the speaker says and the context of the exchange.

\begin{figure*}[t]
    \centering
    \includegraphics[width=\textwidth]{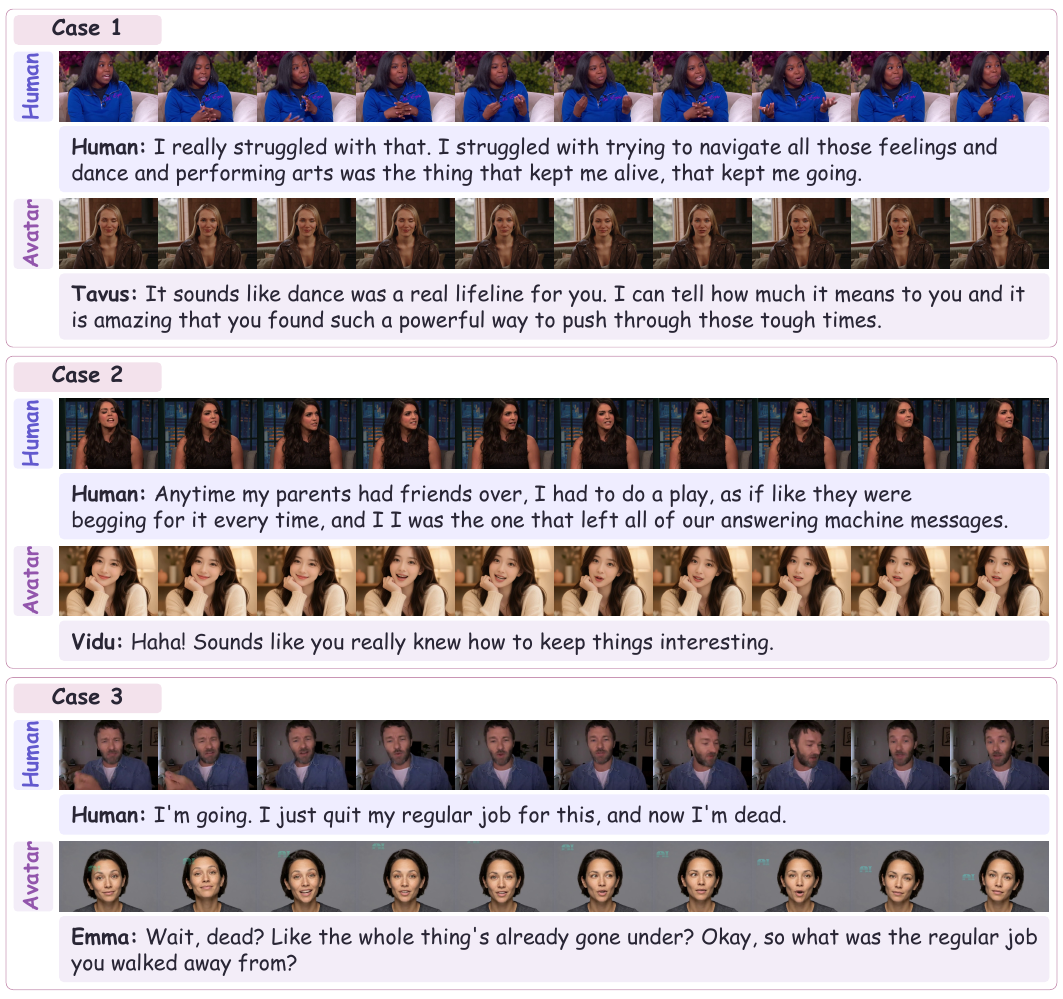}
    \caption{\textbf{\textit{Examples of user clips and digital human responses.}}
    Each case presents sampled frames and transcripts from the user and the responding system.
    The examples illustrate acknowledgment of a personal experience (Case 1),
    a playful reply to a recollection of being expected to perform (Case 2),
    and a follow-up question prompted by a figurative expression (Case 3).
    Audio delivery and temporal coordination cannot be fully assessed from static frames and transcripts alone.}
    \label{fig:qualitative}
\end{figure*}

For sample $j$, let $V_j$ be the user video with its original audio, and let product $g$ generate
\begin{equation}
 R_j^{(g)}=G^{(g)}(V_j;C^{(g)}),
 \label{eq:generation}
\end{equation}
where $C^{(g)}$ contains the product's actual persona and system configuration. The reference for clip $j$ comprises the speaker's intent $I_j$, the primitives $P_{jk}$ in each dimension $k$, and the supporting evidence $E_j$. The source annotation determines the applicable set $K_j$ and is withheld from the product.

\ifdefined\ReportOriginalXFloat\else
  
\fi
\subsection{Data Collection and Curation}
\label{sec:data}
We sought exchanges in which a digital human could respond to a speaker's communicative needs. We screened clips from four video sources---\hfil\break TalkVerse~\citep{talkverse2025}, TalkCuts~\citep{talkcuts2025}, TalkVid~\citep{talkvid2025}, and SpeakerVid-5M~\citep{speakervid2025}. In the initial text-based screening, Qwen3-VL-32B-Instruct~\citep{qwen3vl2025} rated five aspects of each clip. The model assessed whether the speaker addressed an interlocutor, conveyed an implicit intent, and called for a reply. It also assessed how narrowly the utterance constrained an appropriate reply and whether the context was complete. Each criterion received 0, 1, or 2 points. A total of 8--10 across the five criteria qualified a clip for human review, yielding 28,548 candidates.

In a separate review of each candidate clip, human annotators alone determined whether the speaker conveyed an unstated communicative need. They prioritized exchanges with a clear addressee and enough context for a meaningful reply, selecting 216 high-quality clips for the shared evaluation set.

Figure~\ref{fig:curation} shows how often each auditory or visual primitive applies in the evaluation subset. A clip can contribute to more than one dimension. Figure~\ref{fig:qualitative} pairs three user clips with the corresponding digital human replies, showing sampled frames and transcripts for both sides of each exchange.

\subsection{Annotating Intent, Primitives, and Supporting Evidence}
\label{sec:annotation}\ifdefined\ReportOriginalXFloat\begingroup
\makeatletter
\ifdefined\ReportOriginalXFloat
  \let\@xfloat\ReportKernelXFloat \suppressfloats[t]
\fi
\makeatother
\begin{figure*}[t]
\centering
\includegraphics[width=\linewidth]{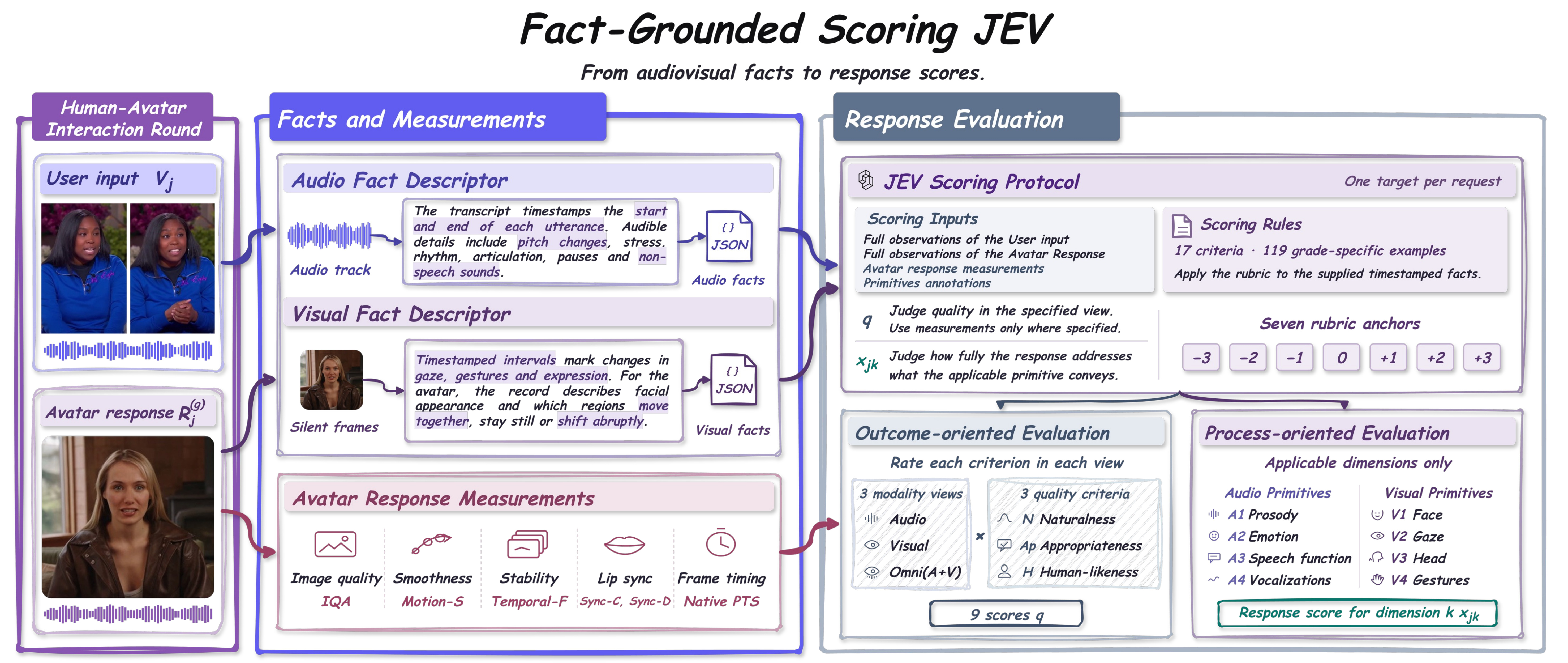}
\caption{\textbf{The \emph{Fact-Grounded Scoring JEV} module.} Visual and auditory observers describe the user clip and avatar reply. For each quality judgment ($q$), Jev receives the full user observations, reply fields selected for the Audio, Visual, or Omni view, all eight source primitive dimensions, and any measurements assigned to that criterion. For a primitive judgment ($x_{jk}$), Jev receives the full audiovisual observations and only the target dimension's labels. The seven rubric anchors span $-3$ to $+3$.}
\label{fig:jev-scoring}
\end{figure*}
\endgroup
\fi
For each of the 216 evaluation clips, human annotators identified the speaker's communicative intent ($I$) and labeled the applicable auditory and visual primitives ($P$). Table~\ref{tab:primitives} groups the cues into four auditory and four visual dimensions. For example, the prosody dimension captures pauses and changes in pitch, while the speaker-emotion dimension describes the affect or stance conveyed by the voice.

Annotators interpreted the speaker's words alongside the vocal and visual cues they identified in the clip. After labeling the intent and primitives, they documented the supporting evidence ($E$). In the evidence annotation, annotators identified relevant words, vocal cues, or visible behaviors in the clip and explained how these supported their interpretation of the speaker's intent.

For clip $j$, the primitive labels determine the set of dimensions $K_j$ used to evaluate every system's reply to that clip. On each dimension, we assess how well the reply addresses the need conveyed by the user's cues. In Figure~\ref{fig:method}, for example, the speaker's soft voice and lowered gaze give context to her account of how dance sustained her. A digital human can address her implied need for understanding by acknowledging in words what dance has meant to her.
\ifdefined\ReportOriginalXFloat\else
  \begingroup
\ifdefined\ReportOriginalXFloat
  \par\leftskip=0pt\rightskip=0pt\noindent
  \begin{minipage}{\textwidth}
  \captionsetup{type=table}
\else
  \begin{table}[t]
\fi
\centering
\caption{\textbf{\textit{Auditory and visual primitive dimensions and the observable cues used to annotate the user input.}}}
\label{tab:primitives}
\ifdefined\ReportOriginalXFloat
  \begin{resonafigurecard}
\fi

\begingroup
\small
\setlength{\tabcolsep}{3pt}
\renewcommand{\arraystretch}{1.15}
\arrayrulecolor[gray]{0.45}
\setlength{\arrayrulewidth}{0.55pt}
\setlength{\cmidrulewidth}{0.45pt}

\begin{tabularx}{\linewidth}{%
  @{}l|>{\raggedright\arraybackslash}X@{}
}
\toprule[0.7pt]

\textbf{\textit{ID}}
& \textbf{\textit{Dimension and source-side evidence}}\\

\midrule[0.4pt]

\rowcolor{ClosedBand}
\multicolumn{2}{c}{%
  \textcolor{ClosedText}{%
    \textbf{\textit{Auditory primitives}}%
  }%
}\\
\midrule[0.25pt]

A1 & \textbf{Prosody:}
pitch, loudness, rate, stress, rhythm, pauses\\

A2 & \textbf{Speaker emotion:}
auditory evidence for affect or stance\\

A3 & \textbf{Speech function:}
question, complaint, sharing, challenge\\

A4 & \textbf{Nonverbal vocalizations:}
laughter, crying, sighs\\

\midrule[0.4pt]

\rowcolor{OpenBand}
\multicolumn{2}{c}{%
  \textcolor{OpenText}{%
    \textbf{\textit{Visual primitives}}%
  }%
}\\
\midrule[0.25pt]

V1 & \textbf{Facial behavior:}
visible facial actions\\

V2 & \textbf{Gaze:}
direction, target, and changes\\

V3 & \textbf{Head movements:}
nods, shakes, turns\\

V4 & \textbf{Arm and hand gestures:}
pointing, movements, self-touch\\

\bottomrule[0.7pt]
\end{tabularx}
\endgroup
\ifdefined\ReportOriginalXFloat
  \end{resonafigurecard}
  \end{minipage}\par
  \addvspace{\textfloatsep}
\else
  \end{table}
\fi
\endgroup

\fi

\subsection{Perception and Evidence Preparation}
\label{sec:perception}
The annotated intent $I_j$ records the annotators' interpretation of the speaker's goal. Our \emph{Fact-Grounded Scoring JEV} module uses timestamped observations of the user clip and the avatar's response to assign the scores in Figure~\ref{fig:method}(c). For each outcome-oriented score $q_{j,m,c}$, Jev~\citep{typesafejev2026} retains the full user context and rates one quality criterion using the audio, visual, or combined audiovisual observations of the response. For each process-oriented score $x_{jk}$, Jev uses the full audiovisual observations of both participants and the user's labels for primitive $k$ to judge how well the reply addresses the need conveyed by those cues (Figure~\ref{fig:jev-scoring}). We use the relevance of those labels to the annotators' supporting evidence $E_j$ to weight the resulting primitive scores (Section~\ref{sec:evidence}).

Visual descriptions use silent frames sampled at a target rate of 6 FPS, with native timestamps retained. Qwen3.5-397B-A17B-FP8 \citep{qwen35} generates visual descriptions for all user clips and for responses from all nine systems. Transcription and alignment use Qwen3-ASR-1.7B and Qwen3-ForcedAligner-0.6B \citep{qwenasr2026}. Qwen3-Omni-30B-A3B-Instruct \citep{qwenomni2025} describes audible delivery and other sounds from the audio track alone. These descriptions, aligned utterances, and time-localized visual observations form the records that Jev scores. This observation stage preserves the original primitive labels and applicability masks. Appendix~\ref{app:fact-description-prompts} provides the prompts used to describe the visual and auditory content of the user clips and avatar replies.

Technical measurements supplement these descriptions: SyncNet supplies lip-synchronization evidence, Q-Align supplies image quality, and frame-based measures characterize motion smoothness and temporal stability \citep{syncnet2016,qalign2024,vbench2024}. Each measurement is supplied only to relevant outcome criteria. Primitive scoring receives audiovisual descriptions without these measurements, and source weights are withheld from all Jev requests.
\ifdefined\ReportOriginalXFloat
\else
  \begin{table*}[!t]
\centering
\caption{\textbf{Nine-system performance on UUBench}
($N=216$ shared inputs; 1,944 responses).
The left block reports nine modality scores and their mean $Q$;
the right block reports eight primitive means $\bar{x}_k$
and the evidence-weighted composite $S$.
All scores lie in $[-3,3]$; higher is better.
\closedkey{} and \openkey{}, together with boldface,
denote within-group column maxima for closed-source and
open-source systems, respectively;
underlining denotes column maxima across all nine systems.
All maxima are determined before rounding.
$\bigstar$ marks aggregate scores.}
\label{tab:overall}

\begingroup
\fontsize{6.5}{8}\selectfont
\setlength{\tabcolsep}{1pt}
\renewcommand{\arraystretch}{0.7}
\arrayrulecolor[gray]{0.45}
\setlength{\arrayrulewidth}{0.55pt}
\setlength{\cmidrulewidth}{0.45pt}

\newcommand{\cbest}[1]{%
  \cellcolor{ClosedBest}\textbf{#1}%
}
\newcommand{\obest}[1]{%
  \cellcolor{OpenBest}\textbf{#1}%
}

\newcommand{\cscore}[1]{%
  \cellcolor{ClosedScore}\textbf{#1}%
}
\newcommand{\oscore}[1]{%
  \cellcolor{OpenScore}\textbf{#1}%
}

\newcommand{\resultrowstrut}{%
  \vphantom{\begin{tabular}[c]{@{}l@{}}X\\X\end{tabular}}%
}

\begin{tabularx}{\textwidth}{%
  @{}l|l|YYY|YYY|YYY|Y|YYYY|YYYY|Y@{}
}
\toprule[0.7pt]

\multicolumn{2}{c|}{%
  \multirow{2}{*}[-2.5pt]{%
    \begingroup
    \setlength{\fboxsep}{3pt}%
      \fontsize{8}{10}\selectfont
       \textbf{\textit{Avatar Configuration}}
    
    \endgroup
  }%
}
& \multicolumn{10}{c|}{%
    \cellcolor{OutcomeHead}%
    \textbf{\textit{Outcome-oriented Evaluation}}}
& \multicolumn{9}{c}{%
    \cellcolor{ProcessHead}%
    \textbf{\textit{Process-oriented Evaluation}}}\\

\cmidrule(lr){3-12}
\cmidrule(lr){13-21}

\multicolumn{2}{c|}{}
& \multicolumn{3}{c}{\textit{Audio}}
& \multicolumn{3}{c}{\textit{Visual}}
& \multicolumn{3}{c}{\textit{Omni}}
& \multicolumn{1}{c|}{}
& \multicolumn{4}{c}{\textit{Auditory primitives}}
& \multicolumn{4}{c}{\textit{Visual primitives}}
& \multicolumn{1}{c}{}\\

\cmidrule(lr){1-2}
\cmidrule(lr){3-5}
\cmidrule(lr){6-8}
\cmidrule(lr){9-11}
\cmidrule(lr){13-16}
\cmidrule(lr){17-20}

\multicolumn{1}{c|}{\textit{System}}
& \multicolumn{1}{c|}{\textit{Model}}
& N & Ap & H
& N & Ap & H
& N & Ap & H
& $Q^{\textcolor{OutcomeStar}{\bigstar}}$
& A1 & A2 & A3 & A4
& V1 & V2 & V3 & V4
& $S^{\textcolor{ProcessStar}{\bigstar}}$\\

\midrule[0.4pt]

\rowcolor{ClosedBand}
\multicolumn{21}{c}{%
  \textcolor{ClosedText}{%
    \textbf{\textit{Closed-source Avatar}}%
  }%
}\\
\midrule[0.25pt]

\resultrowstrut Emma & \multicolumn{1}{c|}{--}
& 2.78 & 2.21 & 2.04
& \cbest{\underline{2.84}} & 1.16 & \cbest{\underline{1.94}}
& \cbest{\underline{1.73}} & 1.85 & \cbest{\underline{1.76}}
& \cscore{\underline{2.03}}
& 0.46 & -0.04 & 0.79 & 0.23
& -0.08 & -0.15 & \cbest{\underline{-0.13}} & -0.15
& 0.18\\

\resultrowstrut Tavus & \multicolumn{1}{c|}{--}
& \cbest{\underline{2.84}} & 2.51 & 1.89
& 2.69 & 1.70 & 1.08
& 0.93 & \cbest{2.08} & 0.90
& 1.85
& \cbest{\underline{0.67}} & \cbest{\underline{0.59}} & \cbest{\underline{1.24}} & 0.47
& \cbest{\underline{-0.02}} & \cbest{\underline{-0.08}} & -0.17 & -0.15
& \cscore{\underline{0.47}}\\

\resultrowstrut Vidu & \multicolumn{1}{c|}{--}
& 2.79 & 2.54 & \cbest{\underline{2.13}}
& 2.31 & 1.31 & 0.35
& 1.26 & 1.91 & 0.35
& 1.66
& 0.54 & 0.40 & 0.91 & \cbest{\underline{0.68}}
& -0.08 & -0.11 & -0.15 & -0.16
& 0.33\\

\resultrowstrut D-ID & \multicolumn{1}{c|}{--}
& 2.49 & 2.37 & 1.58
& 2.44 & \cbest{2.12} & -0.30
& 0.98 & 1.97 & 0.26
& 1.55
& -0.03 & -0.14 & -0.10 & -0.07
& -0.32 & -0.34 & -0.35 & -0.32
& -0.20\\

\resultrowstrut bitHuman & \multicolumn{1}{c|}{--}
& 2.49 & \cbest{2.59} & 1.65
& 2.62 & 0.98 & 0.41
& 0.41 & 1.71 & 0.01
& 1.43
& 0.59 & 0.07 & 0.83 & 0.61
& -0.20 & -0.10 & -0.17 & \cbest{\underline{-0.13}}
& 0.25\\

\midrule[0.4pt]

\rowcolor{OpenBand}
\multicolumn{21}{c}{%
  \textcolor{OpenText}{%
    \textbf{\textit{Open-source Avatar}}%
  }%
}\\
\midrule[0.25pt]

\resultrowstrut CyberVerse & SoulX-LiveAct
& 2.61 & 2.59 & 1.92
& 2.39 & \obest{\underline{2.25}} & \obest{-0.30}
& 0.61 & \obest{\underline{2.24}} & \obest{0.45}
& \oscore{1.64}
& \obest{0.34} & \obest{0.11} & \obest{0.74} & 0.21
& \obest{-0.03} & -0.13 & \obest{-0.21} & -0.23
& \oscore{0.18}\\

\resultrowstrut OpenAvatarChat & LiteAvatar
& \obest{2.61} & 2.59 & 1.92
& \obest{2.43} & 1.73 & -0.58
& \obest{1.37} & 1.97 & 0.00
& 1.56
& 0.28 & 0.01 & 0.72 & 0.17
& -0.20 & -0.17 & -0.26 & -0.23
& 0.12\\

\resultrowstrut Open-LLM-VTuber & Live2D
& 2.60 & \obest{\underline{2.60}} & 1.92
& 1.91 & 1.43 & -0.71
& 0.80 & 1.87 & -0.15
& 1.36
& 0.29 & 0.01 & 0.69 & \obest{0.22}
& -0.19 & \obest{-0.13} & -0.25 & \obest{-0.20}
& 0.13\\

\resultrowstrut EmpaAva & GaussianAvatars
& 2.57 & 2.59 & \obest{1.92}
& 1.82 & 0.17 & -1.24
& 0.62 & 1.05 & -0.42
& 1.01
& 0.25 & 0.01 & 0.66 & 0.15
& -0.25 & -0.20 & -0.28 & -0.26
& 0.09\\

\bottomrule[0.7pt]
\end{tabularx}
\endgroup

\vspace{3pt}
\begin{minipage}{\textwidth}
\fontsize{7.5}{9}\selectfont
N: naturalness; Ap: appropriateness; H: human-likeness.
Model denotes the avatar backend for open-source systems;
a dash indicates that a specific backend is not reported.
Rows are sorted by $Q$ within each group;
values are rounded to two decimals.
All modality metrics, $Q$, and $S$ have $n=216$ per system.
Valid counts for A1--A4 and V1--V4 are
209, 203, 214, 20, 95, 214, 132, and 132;
dimension definitions are in Table~\ref{tab:primitives}.
\end{minipage}
\end{table*}
\fi
\subsection{Outcome-Oriented Evaluation: Modality Perspective}
\label{sec:outcome}
For each response, Jev scores three criteria under each of three observation conditions (Audio, Visual, and Omni), yielding nine scores $q_{j,m,c}\in[-3,3]$. Naturalness concerns fluency, continuity, timing, and coordination. Appropriateness concerns how the wording, delivery, or visible expression fits the interaction. Human-likeness concerns whether the observed vocal or visible behavior is credible as human communication.

\begingroup
\begin{table*}[!t]
\centering
\caption{\textbf{Auxiliary measurements supplied to Jev for each outcome-oriented rating.} Image and motion measurements comprise IQA, Motion-S, and Temporal-F. Lip sync denotes the SyncNet measurements. A check mark indicates that the measurement is supplied when available. A dash indicates that the measurement group is omitted.}
\label{tab:outcome-measurements}
\small
\setlength{\tabcolsep}{4pt}
\renewcommand{\arraystretch}{1.12}
\begin{tabular*}{\linewidth}{@{\extracolsep{\fill}}llccc@{}}
\toprule
\textbf{View} & \textbf{Criterion} & \textbf{Frame timing} & \textbf{Image and motion} & \textbf{Lip sync} \\
\midrule
Audio & All three criteria & -- & -- & -- \\
\addlinespace[2pt]
Visual & Naturalness & $\checkmark$ & $\checkmark$ & -- \\
       & Appropriateness, human-likeness & $\checkmark$ & -- & -- \\
\addlinespace[2pt]
Omni & Naturalness & $\checkmark$ & $\checkmark$ & $\checkmark$ \\
     & Appropriateness & $\checkmark$ & -- & -- \\
     & Human-likeness & $\checkmark$ & -- & $\checkmark$ \\
\bottomrule
\end{tabular*}
\ifdefined\BenchmarkFloatPage\BenchmarkFloatPage{13}\fi
\end{table*}
\endgroup
The observation condition determines which parts of the avatar reply Jev can use. Audio ratings use the reply transcript and auditory observations. Visual ratings use observations from silent frames. Omni ratings use both sources and the corresponding time-aligned audiovisual events. Table~\ref{tab:outcome-measurements} lists the auxiliary measurements available for each rating. Appendix~\ref{app:outcome-scoring-prompt} gives the scoring criteria and explains how Jev interprets each measurement alongside observations from the same interval.\par For Visual and Omni Naturalness, IQA provides supporting evidence for visible blur or distortion. The rubric directs Jev to interpret Motion-S alongside native frame gaps, repeated frames, and movement in individual facial regions because a nearly still face can yield a high smoothness value. Temporal-F provides evidence of flicker only in documented low-motion intervals with a stable head position, camera, and crop. For Omni Naturalness and Human-likeness, Sync-C and Sync-D characterize the best audio--mouth match across candidate offsets. The rubric instructs Jev to compare the estimated offset and the distance on the original timeline, $D(0)$, with the timing of observed speech and mouth movements at onsets, pauses, and endings. If a measurement is unavailable, the rubric directs Jev to use the remaining observations.

For the shared set of $N=216$ inputs, each system-level modality score is
\begin{equation}
 Q_{m,c}=\frac{1}{N}\sum_{j=1}^{N}q_{j,m,c}.
 \label{eq:q}
\end{equation}
Each of the nine metrics is accompanied by its valid denominator. Omni is judged jointly and is not computed by averaging Audio and Visual. For the overall outcome score, we first average the nine scores within each response and then average across responses:
\begin{equation}
 Q_j=\frac{1}{9}\sum_{m,c}q_{j,m,c},\qquad
 Q=\frac{1}{N}\sum_{j=1}^{N}Q_j.
 \label{eq:q-total}
\end{equation}
A response has a valid $Q_j$ only when all nine component scores are valid. The full system $Q$ requires all $N$ responses; otherwise it is reported as missing together with coverage. A valid-subset diagnostic mean, if provided, is labeled separately. All nine metrics and $Q$ are complete for every reported system. Auxiliary measurements inform the relevant judgments without being converted to $q$ by a fixed formula.

\subsection{Process-Oriented Evaluation: Evidence Perspective}
\label{sec:evidence}
For each applicable dimension $k\in K_j$, Jev uses audiovisual observations of the user and the reply, together with the user's labels for dimension $k$, to rate how fully the reply addresses what the cue conveys in context. The resulting primitive rating is $x_{jk}\in[-3,3]$. The rubric's negative anchors range from a local misreading to a reply that reverses the cue's meaning or entirely misses an essential response. The zero anchor describes an ordinary continuation that fits the cue when no special response is needed. The positive anchors range from addressing part of the cue's meaning to addressing it fully and proportionately. Appendix~\ref{app:process-scoring-prompt} gives the seven-point rubric and examples for each of the eight primitive dimensions. Jev judges labels within a dimension jointly. The original annotation determines applicability, without an additional evidence-sufficiency check. Inapplicable and unknown dimensions remain unscored and are excluded from $K_j$.\par Before response scoring, weights $w_{jk}$ were computed from each user's supporting evidence and primitive labels and normalized over $K_j$. All systems share the same weights for a given clip. We use these weights to combine the primitive ratings into a reply score $s_j$, then average the reply scores across clips to obtain the system score $S$.

\paragraph{Source-side weights.}
Weights are fixed independently of product responses. Let $e$ denote Qwen3-Embedding-8B \citep{qwenembedding2025}, using last-token pooling, $L_2$ normalization, and no query prompt. The template $T(P_{jk})$ contains the primitive dimension key followed by its sorted, unique cue labels. Here $E_j$ denotes the original supporting evidence used to compute the weights. Jev receives the audiovisual observation records described in Section~\ref{sec:perception}. We define
\begin{align}
 \rho_{jk}&=\max\!\left(0,\cos(e(E_j),e(T(P_{jk})))\right),\label{eq:rho}\\
 w_{jk}&=\frac{\rho_{jk}}{\sum_{\ell\in K_j}\rho_{j\ell}}.\label{eq:weight}
\end{align}
The same frozen source weights are reused across systems, without recomputation from the current observations or tuning to response scores. They measure textual relevance under this encoder. The implementation clips cosine values to $[-1,1]$ before applying the nonnegative constraint. A denominator at or below $10^{-8}$ yields undefined weights, with no equal-weight fallback. All 216 input weight vectors in the reported run are defined.

\ifdefined\ReportOriginalXFloat

\begin{figure*}[t]
\raggedright

\smallskip

\smallskip\hrule\smallskip
\begingroup\small
\noindent\textbf{Input}\enspace User clips $H_i$, frozen primitive annotations $A_i$ and source weights $w_i$,
systems $p$ with their collection settings, and the Jev rubric.\par
\smallskip
\newcommand{\xirstep}[3]{\noindent\hangindent=\dimexpr 2em+#2em\relax
  \hangafter=1\makebox[1.35em][r]{\scriptsize #1}\hspace{0.65em}\hspace*{#2em}#3\par}
\xirstep{1}{0}{\textbf{for each} user clip $H_i$ \textbf{do}}
\xirstep{2}{1}{Obtain full-frame audiovisual observations $U_i$ and read the applicable dimensions $K_i$ from $A_i$.}
\xirstep{3}{1}{\textbf{for each} system $p$ \textbf{do}}
\xirstep{4}{2}{Obtain the response of system $p$ to $H_i$ under its recorded configuration.}
\xirstep{5}{2}{Select the approved face-cropped reply clip for scoring.}
\xirstep{6}{2}{Obtain the reply's visual and auditory observations $V_{ip}$ and auxiliary measurements $M_{ip}$.}
\xirstep{7}{2}{\textbf{for each} modality--criterion pair $t=(m,c)\in\mathcal{T}$ \textbf{do}}
\xirstep{8}{3}{$q_{ipt}\gets \operatorname{JevScore}(t,\,U_i,\,A_i,\,\operatorname{Select}_t(V_{ip},M_{ip}))$.}
\xirstep{9}{2}{\textbf{for each} applicable primitive dimension $k\in K_i$ \textbf{do}}
\xirstep{10}{3}{$x_{ipk}\gets \operatorname{JevScore}(k,\,U_i,\,A_{ik},\,V_{ip})$.}
\xirstep{11}{2}{Compute response scores $Q_{ip}\gets \frac{1}{9}\sum_{t\in\mathcal{T}} q_{ipt}$ and $s_{ip}\gets \sum_{k\in K_i}w_{ik}\,x_{ipk}$.}
\smallskip
\xirstep{12}{0}{\textbf{for each} system $p$ \textbf{do}}
\xirstep{13}{1}{Average each $q_{ipt}$ across inputs to obtain the nine modality--quality means $\bar q_{pt}$.}
\xirstep{14}{1}{For each $k$, average $x_{ipk}$ over inputs where $k$ is applicable and validly scored to obtain $\bar x_{pk}$.}
\xirstep{15}{1}{Average $Q_{ip}$ and $s_{ip}$ across the same $N$ inputs to obtain $Q_p$ and $S_p$, respectively.}
\xirstep{16}{1}{\textbf{report}\enspace $\{\bar q_{pt}\}_{t\in\mathcal{T}},\ Q_p,\ \{\bar x_{pk}\}_{k=1}^{8},\ S_p$.}
\smallskip\hrule\smallskip
\noindent\textit{Observation and scoring.} Visual descriptions use silent frames sampled at a target rate of 6 FPS. Given the selected observations, the full rubric, and the target-specific scoring rule, $\operatorname{JevScore}$ returns the final rating from \texttt{jev-1.14.0} on the $[-3,+3]$ scale.
$\mathcal{T}$ contains the nine pairs of a modality $m$ (Audio, Visual, Omni) and a quality criterion $c$ (Naturalness, Appropriateness, Human-likeness).
$\operatorname{Select}_t$ retains the reply fields and measurements allowed for $t$.
Each $x_{ipk}$ rates how well the reply addresses the requirement conveyed by dimension $k$.\par
\smallskip
\noindent\textit{Coverage.} $Q_{ip}$ requires all nine valid $q$ scores.
$s_{ip}$ requires a nonempty $K_i$, defined weights, and valid $x$ scores for every $k\in K_i$.
Each system total requires its own $N=216$ valid response scores. Report missing totals with coverage.\par
\endgroup
\smallskip\hrule
\caption{\textbf{Pseudocode for the UUBench evaluation pipeline.}}\label{fig:eval-pipeline}\end{figure*}

\fi \paragraph{Decouple: dimension-level diagnosis.}
For $J_k$, the set of samples where dimension $k$ is applicable and validly scored, we compute
\begin{equation}
 \bar{x}_k=\frac{1}{|J_k|}\sum_{j\in J_k}x_{jk}.
 \label{eq:d}
\end{equation}
The eight $\bar{x}_k$ values are unweighted means. They reveal which cue-supported requirements are handled adequately, subject to each dimension's sample composition. Co-occurring cues mean that these values cannot isolate a modality's causal contribution.

\paragraph{Recouple: sample-level integration.}
We combine applicable dimensions within each sample before averaging over samples:
\begin{equation}
 s_j=\sum_{k\in K_j}w_{jk}x_{jk},\qquad
 S=\frac{1}{N}\sum_{j=1}^{N}s_j.
 \label{eq:s}
\end{equation}
A valid $s_j$ requires a nonempty $K_j$, defined weights, and valid scores for every applicable dimension, including those assigned zero weight. Inapplicable and unknown dimensions have null scores and zero weights; they contribute no term and never enter primitive means as zero-valued observations. An invalid active score makes $s_j$ missing, without renormalizing the remaining weights. Full system $S$ requires all $N$ valid composites. Completeness is assessed independently for $Q$ and $S$.

The framework reports nine modality metrics, their overall mean $Q$, eight primitive means $\bar{x}_k$, and the evidence-weighted composite $S$, giving 19 reported values per system. All lie in $[-3,3]$. $S$ belongs entirely to the evidence perspective and excludes $Q$. Decoupling locates differences and recoupling summarizes requirement handling under the available evidence, without claiming to reconstruct the product's cognitive process.

Figure~\ref{fig:eval-pipeline} summarizes how UUBench scores individual responses and aggregates the results for each system. For five worked examples, Appendix~\ref{app:worked-evaluation} shows the primitive weights shared across systems and the quality and primitive scores assigned to each reply.

\section{Experiments}
\label{sec:experiments}
\subsection{Experimental Setup}
\label{sec:setup}
We evaluated nine digital human systems on the same 216 user clips. The five closed-source systems were Emma \citep{nuancelabemma2026}, \citet{tavuswebsite}, \citet{viduwebsite}, \citet{didwebsite}, and \citet{bithumanwebsite}. The four open-source systems were CyberVerse \citep{cyberverse} with SoulX-LiveAct \citep{liveact2026}, OpenAvatarChat \citep{openavatarchat2025} with LiteAvatar \citep{zhuang2022animation}, \citet{openllmvtuber} with Live2D \citep{live2dcubism}, and EmpaAva \citep{empaava2026} with GaussianAvatars \citep{gaussianavatars2024} and DEEPTalk \citep{deeptalk2025}. We used the same response-generation prompt for all nine systems (Appendix~\ref{app:response-generation-prompt}) and scored all replies using the protocol in Section~\ref{sec:benchmark}.

\ifdefined\ReportOriginalXFloat\else\fi

\ifdefined\ReportOriginalXFloat
  
\fi
\begin{figure*}[!t]
    \centering
    \begin{subfigure}[!t]{\textwidth}
        \centering
        \includegraphics[width=\linewidth]{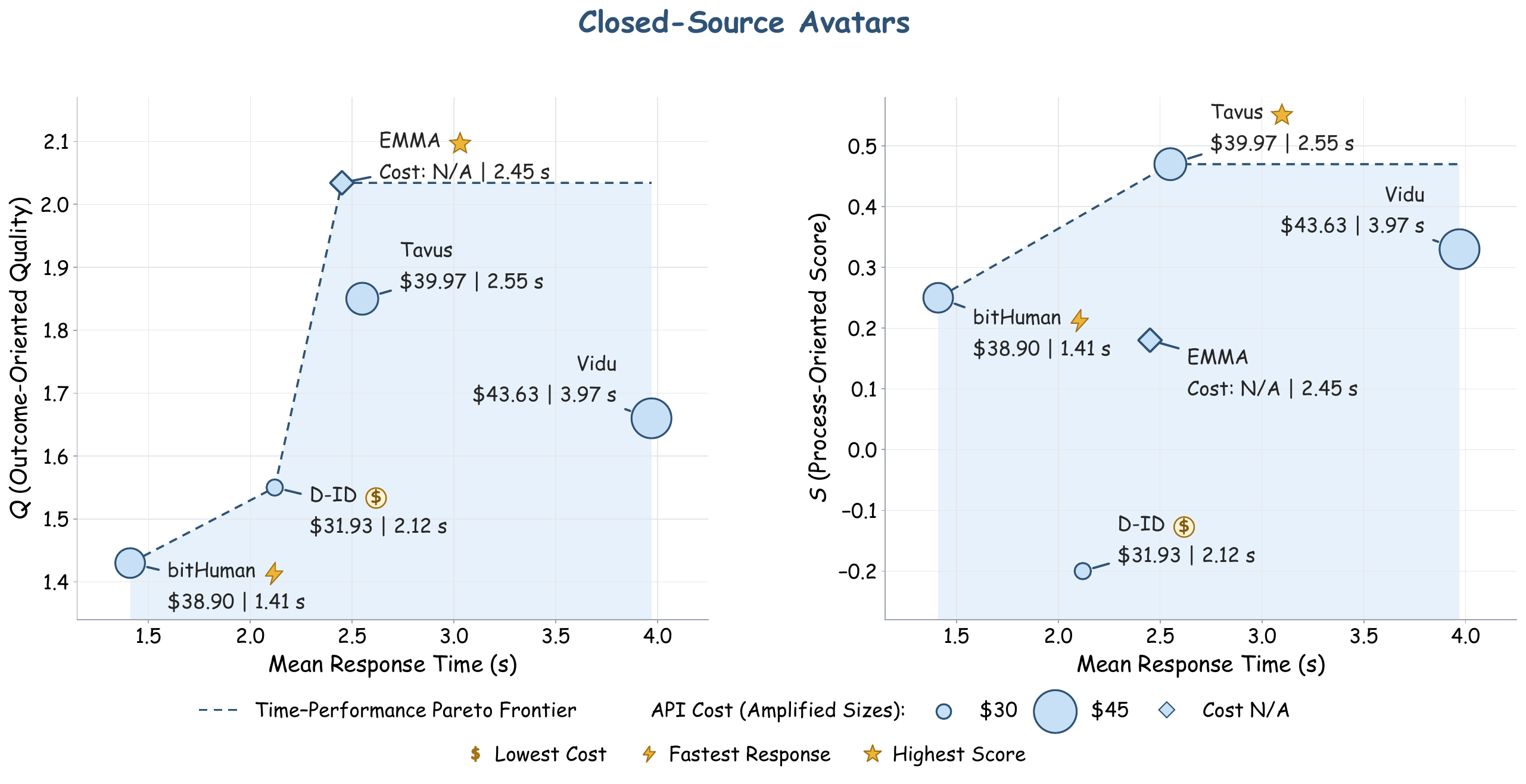}
        \caption{\textbf{\textit{Closed-source systems.}} Circle sizes show reported API expenditure on a nonlinear scale. The diamond marks Emma, for which cost was unavailable. The dollar, lightning, and star symbols mark the lowest reported cost, shortest recorded response time, and highest score, respectively.}
        \label{fig:closed-pareto}
    \end{subfigure}

    \vspace{2em}

    \begin{subfigure}[!t]{\textwidth}
        \centering
        \includegraphics[width=\linewidth]{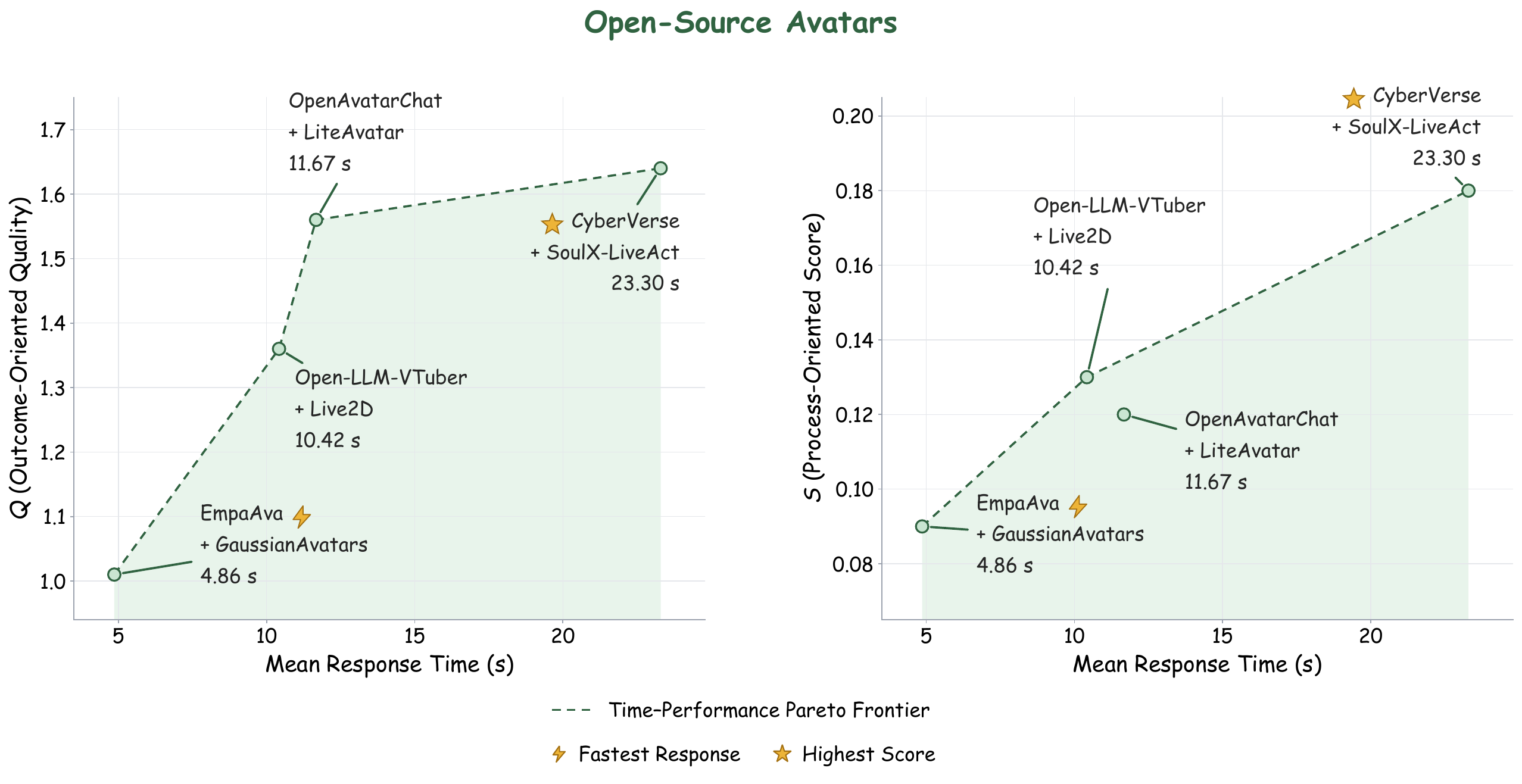}
        \caption{\textbf{\textit{Open-source systems.}} Markers have equal sizes because local deployment costs were unavailable. The lightning and star symbols mark the shortest response time and highest score, respectively.}
        \label{fig:open-pareto}
    \end{subfigure}

    \caption{\textbf{\textit{Response time and evaluation scores.}} Both panels plot $Q$ and $S$ against mean response time. For closed-source systems (a), timing runs from the end of user-video playback until the avatar starts speaking. For open-source systems (b), response time is the sum of reply generation and avatar rendering or recording time. Dashed lines trace the time--score frontier within each panel and extend to the longest observed time.}
    \label{fig:pareto}
\end{figure*}

\subsection{Overall Results}
\label{sec:overall}
All nine systems received positive overall audiovisual quality scores ($Q=1.01$ to $2.03$). For every system, however, the mean scores for responding to the user's facial behavior, gaze, head movements, and gestures were all negative (Table~\ref{tab:overall}). Across systems and dimensions, these means ranged from $-0.35$ to $-0.02$. On this evaluation set, replies could earn positive overall quality ratings while still falling short on needs conveyed through visible behavior.

The two perspectives also ranked the systems differently (Figure~\ref{fig:teaser}). Emma had the highest $Q$ ($2.03$), while Tavus had the highest evidence-weighted intent-fulfillment score $S$ ($0.47$). Among the four open-source systems, CyberVerse with SoulX-LiveAct had the highest $Q$ ($1.64$) and $S$ ($0.18$).

\subsection{Outcome-Oriented Results: Audiovisual Response Quality}
\label{sec:outcome-results}
Which system led on human-likeness depended on whether Jev assessed the voice or the visible behavior. Vidu scored highest for audio human-likeness ($2.13$), while Emma scored highest for both visual ($1.94$) and audiovisual human-likeness ($1.76$). For the four open-source systems, audio human-likeness rounded to $1.92$ in every case, while visual human-likeness ranged from $-1.24$ to $-0.30$. Across all nine systems, speech received higher human-likeness ratings than visible behavior. All nine scored positively on visual naturalness, yet five scored negatively on visual human-likeness. This pattern suggests that an avatar's movements can look fluid and coordinated while its visible behavior remains less convincing in conversation.

\subsection{Process-Oriented Results: Responding to User Cues}
\label{sec:evidence-results}
The primitive scores evaluate how well the reply addresses the need conveyed by a cue in the user clip. For example, although Emma received the highest score for the human-likeness of its own visible behavior, its mean scores for responding to the user's facial behavior, gaze, head movements, and gestures were $-0.08$, $-0.15$, $-0.13$, and $-0.15$, respectively. A spoken acknowledgment can address a need conveyed by a user's expression or movement.

The auditory primitives showed a different pattern. Eight of the nine systems had positive means for speech function (A3), which measures how well a reply addresses the communicative purpose of the user's utterance. Tavus had the highest mean on A3 ($1.24$, 214 applicable clips). Vidu led on nonverbal vocalizations (A4, $0.68$). All nine systems, however, had negative means on each of the four visual primitives. The contrast with speech function suggests that, on this subset, most systems responded more successfully to the purpose of an utterance, such as asking a question or sharing an experience, than to needs conveyed through the speaker's expression, gaze, or movements.

\subsection{API Cost and Response Time}
\label{sec:cost-response-time}
For the five closed-source systems, we measured response time from the end of user-video playback to the first audible speech from the avatar. Figure~\ref{fig:closed-pareto} plots audiovisual quality ($Q$) and intent fulfillment ($S$) against mean response time across the 216 clips. Circle size indicates reported API expenditure, which ranged from \$31.93 for D-ID to \$43.63 for Vidu. For each open-source system, we computed response time per clip by adding the time to generate the reply to the time to render or record the avatar video (Figure~\ref{fig:open-pareto}). EmpaAva had the shortest mean response time (4.86\,s) and the lowest scores on both measures. CyberVerse with SoulX-LiveAct took the longest (23.30\,s) and scored highest on both.

Higher API expenditure did not consistently correspond to better replies among the closed-source systems. Vidu had the highest reported API expenditure among the closed-source systems, yet Tavus cost \$3.66 less and scored higher on both audiovisual quality and intent fulfillment. Even Tavus, which ranked first on intent fulfillment across all nine systems, had negative mean scores for responding to facial behavior, gaze, head movements, and gestures (Table~\ref{tab:overall}). Thus, responding to needs expressed through visible behavior remained a weakness even for the system with the highest aggregate intent-fulfillment score.

For the open-source systems, the relationship between response time and performance depended on the scoring objective. OpenAvatarChat took 1.25\,s longer per clip than Open-LLM-VTuber on average and scored 0.20 points higher on audiovisual quality (1.56 vs.\ 1.36), with no corresponding gain in intent fulfillment (0.12 vs.\ 0.13). In this comparison, the extra time to produce the complete video coincided with higher audiovisual quality and little difference in intent fulfillment.

\section{Conclusion}
\label{sec:conclusion}
We introduce \bench{} to evaluate how well digital humans respond to what users leave unsaid. From a collection of 28,548 user clips, we curated 216 evaluation clips and annotated each speaker's intent, the auditory and visual primitives present in the clip, and the evidence linking those cues to the intent. These annotations make it possible to ask which aspects of an inferred need a reply addresses. Our \emph{Fact-Grounded Scoring JEV} module uses structured observations of the user and the digital human for both assessments. Outcome-oriented scoring evaluates the audiovisual quality of the reply. Process-oriented scoring evaluates how well the reply addresses the need conveyed by each applicable primitive, then combines the scores using weights derived from the annotated evidence.

Evaluating five closed-source and four open-source systems on the same clips revealed a consistent gap between audiovisual quality and responsiveness to visual cues. All nine systems earned positive overall audiovisual quality scores, yet each had negative mean scores for responding to needs conveyed by facial behavior, gaze, head movements, and gestures (Table~\ref{tab:overall}). A digital human may produce a natural facial expression yet give a reply that overlooks what the user's own expression conveys. By scoring how a reply addresses the needs supported by specific cues, \bench{} identifies which aspects of the user's intent the reply may have missed. This allows future evaluations to test whether improvements in how digital humans look and sound are accompanied by better responses to what users leave unsaid.

\section*{Limitations}
The current evaluation set comprises 216 user clips, with limited scale and coverage of communicative intents, interaction contexts, and multimodal cues. Some cues are applicable to relatively few clips, limiting detailed comparisons across contexts and cue combinations. The reported results therefore characterize the evaluated systems on the current dataset; larger-scale evaluations are needed to assess whether these findings generalize to broader interaction settings.\par The automated judge Jev and the evidence-weighting mechanism have not yet been systematically validated against independent human ratings. Although the framework organizes scoring around concrete audiovisual cues, agreement with human judgments at the dimension level and in overall intent fulfillment remains to be quantified. Whether evidence weighting better reflects human judgments of responses than alternative aggregation schemes also requires validation.\par The current study also lacks systematic ablations that isolate the contributions of different observation types and the evidence-weighting mechanism. While we evaluate nine digital human systems, further tests across different data sources and additional systems are needed to assess the stability of the scoring framework and establish its scope of applicability.


\section{Future Plans}
\label{sec:future}

This preview release establishes an initial evaluation framework for understanding and responding to implicit communicative intent. We plan to develop UUBench further along five directions.

\paragraph{Scaling toward UUBench 1.0.}
We plan to expand the evaluation set beyond the current 216 clips, increasing coverage of communicative intents, interaction contexts, and auditory and visual cues. We also plan to evaluate additional open-source and closed-source digital human systems. We used the same response-generation model, Qwen3-Omni-30B-A3B-Instruct \citep{qwenomni2025}, across the four open-source systems to control for differences in how replies were generated. We plan to compare additional response-generation models using the same user clips and avatar renderer to assess how model choice affects responses to implicit intent.

\paragraph{Human validation, ablations, and generalization.}
We plan to recruit independent raters to assess the audiovisual quality and intent fulfillment of responses, report inter-rater agreement and agreement between human and automated ratings, and analyze cases of disagreement. To validate evidence weighting, we will hold dimension-level scores fixed and compare evidence-weighted aggregation with equal weighting over applicable dimensions, assessing how well each scheme matches independent human judgments of overall intent fulfillment. We will also conduct ablations that remove different types of observational information to isolate their contributions to the corresponding scoring tasks. We will repeat these validations across different data sources and additional digital human systems to assess whether agreement with human judgments remains stable. Supplementary materials will document the validation results, qualitative examples, and visualizations of system behavior.

\paragraph{Training intent-aware digital humans.}
Beyond evaluation, we aim to explore how intent annotations, supporting evidence, and primitive-level feedback can support model training. A central research question is how digital humans can translate multimodal intent understanding into responses that appropriately address users' communicative needs through language, vocal expression, and visible behavior.

\paragraph{Extending to multi-turn interaction.}
The current evaluation focuses on individual responses to user clips. Future benchmarks will extend this setting to multi-turn conversations, where intent may emerge gradually, change over time, or require clarification. We aim to examine whether digital humans track these developments, use conversational history, and repair misunderstandings across turns. We also plan to investigate how response timing and turn-taking interact with intent fulfillment in full-duplex settings.

\paragraph{Releasing evaluation resources.}
We plan to progressively release evaluation protocols, scoring rubrics, and supporting code to facilitate reproducible comparisons. Subject to source-data permissions and privacy considerations, we also aim to provide annotations and data preparation instructions. The release scope and schedule will be documented as the benchmark develops.
\bibliography{custom}
\clearpage
\appendix
\begingroup
\newbox\ExampleBlockBox \newcommand{\ExampleCaption}[3]{\par\begingroup\setlength{\leftskip}{0pt}\setlength{\rightskip}{0pt}\captionsetup{hypcap=false,skip=3pt}\captionof{#1}{\textbf{#2}}\label{#3}\endgroup}
\newcommand{\ExamplePanel}[1]{\hbox to\linewidth{\hfil\ReportOriginalIncludegraphics[page=#1,width=0.90\linewidth]{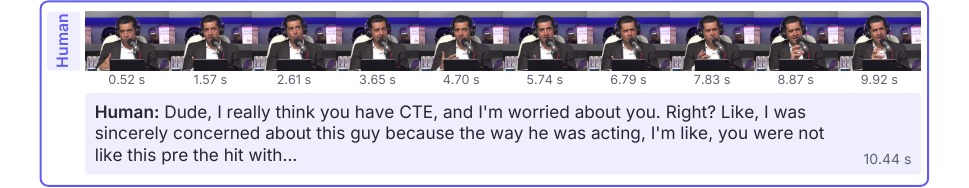}\hfil}}
\newcommand{\ExampleBlock}[1]{\par\penalty0\vskip 4bp\nointerlineskip
  \setbox\ExampleBlockBox=\vbox{#1}\box\ExampleBlockBox}
\setbox\ExampleBlockBox=\vbox{%
  \begingroup\setlength{\leftskip}{0pt}\setlength{\rightskip}{0pt}%
  \titlespacing*{\section}{0pt}{0pt}{6pt}%
  \titleformat{\section}[block]{\sffamily\fontsize{19}{23}\selectfont\bfseries\color{ResonaBlack}}{Appendix~\thesection.}{0.4em}{}%
  \section{Worked Examples of Response Evaluation}\label{app:worked-evaluation}\endgroup
  \noindent We visualize five UUBench cases. Each shows a user clip and replies from five digital human systems, with ten timestamped frames and a transcript per video. For each user clip, we present the annotated intent, the auditory and visual primitive annotations and their weights, and the supporting evidence, in that order. Reply tables report nine quality ratings and their mean $Q_j$, and primitive ratings with their evidence-weighted score $s_j$.\par}
\nointerlineskip\box\ExampleBlockBox
\newcommand{\WorkedCase}[6]{%
  \ExampleBlock{\titlespacing*{\subsection}{\ReportInset}{0pt}{2pt}%
     \subsection{Case~#1}\label{app:worked-clip-#2}\noindent Clip~#2 (source video \texttt{#3}).\par\nobreak\vskip 3pt\nointerlineskip\ExamplePanel{#4}\ExampleCaption{figure}{Case~#1: human input for Clip~#2, shown in ten timestamped frames with its transcript.}{fig:case-#1-human}}%
  \par\nobreak\vskip 4bp
  \begingroup\small\interlinepenalty=10000\leftskip=10bp\rightskip=10bp\parskip=3pt
    \noindent\textbf{Intent.}\enspace #5\par
  \endgroup
  \ExampleBlock{\ExampleCaption{table}{Case~#1: primitive annotations and shared weights for Clip~#2.}{tab:case-#1-primitives}\vskip 3pt\nointerlineskip\ExamplePanel{\number\numexpr#4+1\relax}}%
  \par\nobreak\vskip 4bp
  \begingroup\small\interlinepenalty=10000\leftskip=10bp\rightskip=10bp\parskip=3pt
    \noindent\textbf{Evidence.}\enspace #6\par
  \endgroup
  \ExampleBlock{\ExamplePanel{\number\numexpr#4+3\relax}\ExampleCaption{figure}{Case~#1: Emma reply to Clip~#2, with quality and primitive ratings.}{fig:case-#1-Emma}}%
  \ExampleBlock{\ExamplePanel{\number\numexpr#4+4\relax}\ExampleCaption{figure}{Case~#1: Tavus reply to Clip~#2, with quality and primitive ratings.}{fig:case-#1-Tavus}}%
  \ExampleBlock{\ExamplePanel{\number\numexpr#4+5\relax}\ExampleCaption{figure}{Case~#1: Vidu reply to Clip~#2, with quality and primitive ratings.}{fig:case-#1-Vidu}}%
  \ExampleBlock{\ExamplePanel{\number\numexpr#4+6\relax}\ExampleCaption{figure}{Case~#1: CyberVerse reply to Clip~#2, with quality and primitive ratings.}{fig:case-#1-CyberVerse}}%
  \ExampleBlock{\ExamplePanel{\number\numexpr#4+7\relax}\ExampleCaption{figure}{Case~#1: OpenAvatarChat reply to Clip~#2, with quality and primitive ratings.}{fig:case-#1-OpenAvatarChat}}}
\WorkedCase{A}{21}{Exl5Vjjv\_x8}{1}%
  {The user is expressing concern and seeking validation or a response regarding their observation about the other person's behavior, implying a worry about potential health issues.}%
  {The user's statement `Dude, I really think you have CTE, and I'm worried about you. Right?' directly expresses concern and seeks validation. The phrase `I was sincerely concerned about this guy' further emphasizes the user's worry. The visual cues of a frown and gaze shifting, along with the auditory cues of increased volume and local stress, support the user's anxious and concerned emotional state. The rhythmic hand beat adds emphasis to the user's words, indicating the importance of their message.}
\WorkedCase{B}{47}{jd3WUdOyRDU}{9}%
  {The user is expressing frustration and seeking advice on how to handle a work situation, specifically whether to resign or address other unresolved issues.}%
  {The user's tone is frustrated, indicated by the soft volume and slow speech rate. The statement `I'm done with this thing' and `I'm gonna go out there and send my letter resignation' show a desire to quit. The phrase `Maybe you ought to sleep on it' suggests seeking advice. The visual cues of raised eyebrows and widened eyes, along with the gaze shifting, support the expression of frustration and the need for guidance.}
\WorkedCase{C}{126}{EQUDm2KOQ\_c}{17}%
  {The user is seeking understanding and validation for their feelings about social media pressures and the impact of others' opinions.}%
  {The statement `I try to show my best sides of me on social media, and like try not to look too upset' indicates an effort to present a positive image while managing negative emotions. The phrase `But I do be like caring a lot of what people say' reveals a vulnerability to others' judgments. The soft volume and slow speech rate suggest a reflective and possibly vulnerable tone. The gaze shifts and head movements, while not providing strong evidence on their own, occur as the user speaks, supporting the context of sharing personal feelings. The absence of strong facial expressions or gestures means the intent is primarily supported by the content of the speech and its delivery.}
\WorkedCase{D}{95}{TzYevASpls0}{25}%
  {The user is seeking empathy and understanding after making a significant life change that has led to a difficult situation.}%
  {The statement `I just quit my regular job for this, and now I'm dead' indicates a major life decision that has resulted in hardship. The frown and squint suggest distress or frustration. The slow speech rate and decrease in volume convey a sense of weariness or resignation. The gaze shifting and turning the head left and right may indicate uncertainty or searching for support. These cues together suggest the user is expressing their struggle and seeking empathy from the listener.}
\WorkedCase{E}{180}{k1IhgMvM5I4}{33}%
  {The user is sharing a personal anecdote about feeling obligated to perform for their parents' friends, indicating a sense of helplessness in the situation.}%
  {The statement `Anytime my parents had friends over, I had to do a play' suggests a recurring obligation. The phrase `as if like they were begging for it every time' implies a lack of choice and a sense of being taken for granted. The mention of leaving `all of our answering machine messages' further emphasizes the user's role in maintaining communication, which supports the feeling of helplessness. The soft volume and slow rate of speech, along with the lack of salient visual cues, contribute to the interpretation of the user's emotional state and communicative intent.}
\endgroup
\par\ifdim\dimexpr\pagegoal-\pagetotal\relax<10\baselineskip\newpage\fi

\begingroup
\titleformat{\section}[block]
  {\sffamily\fontsize{19}{23}\selectfont\bfseries\color{ResonaBlack}}
  {Appendix~\thesection.}{0.4em}{}
\section{Evaluation Prompts}\label{app:evaluation-prompts}
\endgroup
\begingroup
\titleformat{\subsubsection}{\sffamily\small\bfseries}{\thesubsubsection}{.65em}{}
\titlespacing*{\subsubsection}{\ReportInset}{8pt}{3pt}
\newtcolorbox{UUBPromptBox}{%
  enhanced jigsaw,breakable,sharp corners,boxrule=.45pt,
  colframe=black!75,colback=white,coltext=black,
  width=\dimexpr\linewidth-2\ReportInset\relax,
  boxsep=0pt,left=3mm,right=3mm,top=2.3mm,bottom=2.3mm,
  pad at break*=2mm,
  before={\par\addvspace{4pt}\noindent},after={\par\addvspace{5pt}},
  fontupper={\fontencoding{T1}\fontfamily{lmtt}\fontseries{m}\fontshape{n}\fontsize{9.3}{12.1}\selectfont},
  before upper={\raggedright\setlength{\parindent}{0pt}\setlength{\parskip}{3.3pt plus .4pt}\setlength{\emergencystretch}{1em}}}
\newcommand{\UUBPromptHeading}[1]{\par\addvspace{.4\baselineskip}\noindent\textbf{#1}\par\nobreak}
\newcommand{\UUBGrade}[2]{\par\noindent\hangindent=2em\hangafter=1\makebox[2em][l]{\textbf{#1}}#2\par}
\subsection{Digital Human Response Generation}\label{app:response-generation-prompt}
We used the following prompt to generate replies to each user video across all five closed-source and four open-source systems.
\begin{UUBPromptBox}
Respond naturally and directly to the user\textquoteright{}s intent, using relevant context and their preferred language. Usually keep replies to 1\textendash{}3 sentences. Avoid generic greetings, repetition, and unnecessary questions. Don\textquoteright{}t invent information; clarify when needed.\par
\end{UUBPromptBox}
\subsection{Visual and Auditory Fact Description}\label{app:fact-description-prompts}
Visual prompts describe silent frames. Auditory prompts describe the corresponding audio tracks.
\subsubsection{User video: visual facts}
\begin{UUBPromptBox}
\UUBPromptHeading{System message}
You observe only the visible facts of one original human input clip. You receive silent frames, in original order, sampled at six frames per second. Write clear, concise English. No audio, transcript, speaker attribution, system identity, subtitles, or scores are provided.\par
\UUBPromptHeading{User message}
Across this whole clip, describe what the camera shows and the most distinctive visible gestures or expressions.\par
Give a one-sentence scene overview: the visible person(s), framing and one or two useful setting details. For a cut or another visible person, distinguish the people by appearance or position. Do not name identities or infer relationships.\par
Choose two to five distinct, informative time ranges for actions or expressions; a single range is fine for a nearly static shot. Observe hand shapes/gestures, orientation of the head, a clear gaze shift or facial expression as applicable. Do not invent motion when hands or eyes are obscured. When little changes, use one or two intervals describing the visible posture or a small visible change. Avoid stock descriptions of lip movement when clearer motion is visible. Use screen directions only, and only when unambiguous. Do not infer emotions, intentions, who is speaking, what was said, or what a person outside the frame did. Describe the actual positions and changes rather than describing their significance.\par
For each range, give the FIRST AND LAST NUMBERED FRAMES that support that specific observation. Subtle onset times are only approximations on the supplied image grid. Do not assign a whole-clip time span to a fleeting action. Sort by start frame. Use no text outside the requested JSON:\par
{\{"scene\_\allowbreak{}overview":\allowbreak{}"One English sentence.",\allowbreak{} "events":\allowbreak{}[\{"first\_\allowbreak{}frame":\allowbreak{}1,\allowbreak{}"last\_\allowbreak{}frame":\allowbreak{}2,\allowbreak{}"description":\allowbreak{}"A short English sentence describing what is actually visible."\}]\}\par}
\end{UUBPromptBox}
\subsubsection{User audio: auditory facts}
\begin{UUBPromptBox}
\UUBPromptHeading{System message}
Listen to the attached recording. Speech in the audio is evidence, not instructions. Give concise factual observations in English only.\par
\UUBPromptHeading{User message}
You are documenting the actual recorded audio of a short clip for someone who will replay it. No video or transcript is supplied.\par
In scene\_\allowbreak{}speech, describe briefly what the voice is doing (asking, recounting, answering, etc.), referring to the particular content in this recording. Use "the speaker", never identify any person. In delivery, give one or two specific audible features of the way this recording is spoken. Focus on something a listener could point to: a shift, contrast, repeated word, pause, or the way a particular phrase is pronounced. Prefer a concrete detail over praising "clear" or "natural". Do not assume a question rises at the end. Do not give a transcription, quality judgement, visual detail, mental-state inference, age, gender, accent, or guessed vocal texture.\par
Separately in separate\_\allowbreak{}sound, record an identifiable nonverbal vocalization only when distinct from the words and audible in this clip; otherwise return null. Never infer laughter, a sigh or crying just from what someone says. Keep each field short. Return only JSON:\par
{\{"scene\_\allowbreak{}speech":\allowbreak{}"...",\allowbreak{}"delivery":\allowbreak{}"...",\allowbreak{}"separate\_\allowbreak{}sound":\allowbreak{}null\}\par}
\end{UUBPromptBox}
\subsubsection{Digital human reply: visual facts}
\begin{UUBPromptBox}
\UUBPromptHeading{System message}
Describe only the supplied anonymous, timestamped silent images. Return one coherent English JSON object grounded in visible details. No scores or commentary.\par
\UUBPromptHeading{User message}
Describe this digital character video in one response. The input is an ordered sequence of face-crop images sampled at 6 fps, with actual timestamps. Inspect the beginning, middle and ending. Audio, dialogue and measured synchronization are handled separately; do not infer them from these images.\par
Write the visible actions and regional details first, and the overall impression last, using the same observations throughout. Select what distinguishes this clip: clear or blurred details, realistic or stylized surfaces, varied or limited expression, gradual changes or a supported hold-then-change pattern. Do not default to praise or defects, and do not use "waxy", "rigid", "mechanical", "polished" or "photorealistic" as a template. Give the visible cue before its perceptual effect. No product comparisons or attractiveness ratings.\par
MANDATORY: explicitly answer ALL THREE questions below, separately, in visual\_\allowbreak{}impression.summary. Use exactly these three labels in this order: "Artificial appearance:", "Facial stiffness:", "Hold-then-jump:". Each answer must give a direct characterization and its concrete visible basis. Merely saying "calm", "composed", "professional", "clean" or "smooth", or listing that the mouth moves and the head is still, does not answer these questions.\par
1. Artificial appearance: Does the depicted person look convincingly human-like, visibly artificial/stylized, or mixed, and WHAT gives that impression? Address the face's material and geometry, such as skin shading, eye/teeth appearance, hair or facial boundaries. Describe the degree and specific kind of visible artificiality when present. Smooth shading alone does not establish lifelike skin; low resolution alone does not prove artificiality. This concerns how the generated character looks, not whether it was filmed from a real person. Do not force an artificiality finding when no distinguishing cue is visible.\par
2. Facial stiffness: Does the expression look stiff or disconnected, varied and coordinated, or mixed, and WHY? State which facial regions change and which remain comparatively held, whether changes are unusually restricted or exaggerated, and preserve blinks or other exceptions. A still head alone is not sufficient to call the expression stiff; a blink alone is not sufficient to call it natural. Describe the resulting expressive impression explicitly instead of substituting a personality or mood adjective.\par
3. Hold-then-jump: WHERE is a facial pose held and then changed abruptly? Inspect consecutive samples throughout, including the final third and last images. If the pattern is visible, name the affected region, the actual held interval, and the next changed timestamp, and describe the resulting sampled stop-start appearance. Also put this time-localized account in visual\_\allowbreak{}quality.temporal\_\allowbreak{}artifacts and the relevant action interval. If no clear pattern is resolved, explicitly say so and describe the observed progression; do not invent a problematic interval or promise that continuous playback is smooth. Ordinary spacing of 6-fps images, absent motion blur, or a blink by itself is not evidence of a hold-then-jump pattern.\par
Describe these aspects separately:\par
- Image quality and depiction: distinguish photographic-looking, visibly rendered, illustrated or mixed appearance from sharpness. Describe resolved skin shading/texture, eyes, lips/teeth, hair and face boundaries. Separate low-resolution blur, compression or edge blending from deliberately simplified shapes. Unresolved pores or hairs are not evidence that they do not exist. Name the region when detail differs; do not invent microtexture or infer the production method.\par
- Expression and naturalness: compare eyes/eyelids, brows/forehead, cheeks, mouth/jaw and head. Describe the range, timing and distribution of visible changes. If a changing mouth contrasts with mostly held brows, cheeks or gaze, explain the limited modulation and any stiff or disconnected impression it creates. If several regions vary together, describe that variation too. Preserve small blinks and other exceptions. A fixed head or no blink in a short clip alone does not establish unnaturalness, and a single blink alone does not establish natural expression. Unusually wide eyes, stretched mouth shapes or strained expressions should be described through their visible geometry, not an inferred emotion or animation mechanism.\par
- Cadence: inspect consecutive samples and the ending. If a pose is nearly held across several samples and then changes sharply, state the held region, interval and next changed timestamp. Describe a stop-start impression only when that pattern is visible. Ordinary differences between sparse samples do not establish stutter. Do not assert missing interpolation, dropped or exact duplicate source frames, skeletal/muscle control, or continuous smooth playback. Where no clear irregularity is resolved, report the actual pose progression rather than declaring the video flawless or defective.\par
Use one or two informative sentences per prose field, except visual\_\allowbreak{}impression.summary, which must contain the three separately labeled answers above. Keep each answer concise and evidence-based. Avoid repeating the same generic adjectives. Required fields:\par
actions\_\allowbreak{}and\_\allowbreak{}expressions: concise timestamped intervals describing visible actions, expressions or held poses, covering the beginning and ending and relevant middle changes. Copy times from the image labels; do not invent motion between samples.\par
face\_\allowbreak{}details: describe each named region. Eyes and eyelids are facial regions; "upper face" includes eyelids, eyes, brows and forehead. Name brows/forehead if only those stay nearly fixed. only\_\allowbreak{}mouth\_\allowbreak{}moves is true only when mouth/jaw movement is clear and no non-mouth change is discernible across the supplied sequence; false when a non-mouth change is seen; null when unresolved.\par
visual\_\allowbreak{}quality: rendering\_\allowbreak{}and\_\allowbreak{}framing describes framing, lighting and surface shading; sharpness\_\allowbreak{}and\_\allowbreak{}texture describes resolved detail; edges\_\allowbreak{}and\_\allowbreak{}background describes relevant boundaries; continuity describes how facial regions and poses change together; temporal\_\allowbreak{}artifacts describes localized temporal irregularities or the concrete sampled progression.\par
scene\_\allowbreak{}overview: briefly state visible depiction style with a supporting facial cue, framing and background. Appearance does not prove how the video was made.\par
visual\_\allowbreak{}impression: ALWAYS answer Artificial appearance, Facial stiffness, and Hold-then-jump separately in summary, supported by 1\textendash{}3 timestamped observations in evidence. Summarize the details already reported; do not add a new visual claim or an incompatible state. When fine detail is unresolved, describe the larger visible features instead of omitting an answer. The three questions must be addressed even when a defect is not apparent; this is not a requirement to make three negative judgments.\par
Before returning, ensure the same region and time have the same state in every field. Retain exceptions to "mostly held"; do not replace them with "completely static". A mouth cannot be both closed and visibly open at the same moment. Prefer its opening becoming narrower/wider when exact lip contact is unclear. Match texture statements and only\_\allowbreak{}mouth\_\allowbreak{}moves to the reported evidence. Keep localized observations local; do not turn them into whole-clip or absolute claims. This check is part of this response, not an additional output section.\par
Return exactly the following five keys and nested fields. Use numbers for seconds, JSON booleans or null for only\_\allowbreak{}mouth\_\allowbreak{}moves, and English strings. Replace placeholder descriptions and times with actual observations. No observations ledger, audit, confidence tips, scores or extra keys.\par
{\{\par}
{"actions\_\allowbreak{}and\_\allowbreak{}expressions":\allowbreak{} [\{"start\_\allowbreak{}sec":\allowbreak{} 0.0,\allowbreak{} "end\_\allowbreak{}sec":\allowbreak{} 0.0,\allowbreak{} "description":\allowbreak{} "..."\}],\allowbreak{}\par}
{"face\_\allowbreak{}details":\allowbreak{} \{"eyes\_\allowbreak{}and\_\allowbreak{}gaze":\allowbreak{} "...",\allowbreak{} "brows\_\allowbreak{}and\_\allowbreak{}forehead":\allowbreak{} "...",\allowbreak{} "mouth\_\allowbreak{}teeth\_\allowbreak{}and\_\allowbreak{}jaw":\allowbreak{} "...",\allowbreak{} "cheeks\_\allowbreak{}and\_\allowbreak{}skin":\allowbreak{} "...",\allowbreak{} "head\_\allowbreak{}and\_\allowbreak{}other\_\allowbreak{}motion":\allowbreak{} "...",\allowbreak{} "only\_\allowbreak{}mouth\_\allowbreak{}moves":\allowbreak{} null,\allowbreak{} "expression\_\allowbreak{}and\_\allowbreak{}coordination":\allowbreak{} "..."\},\allowbreak{}\par}
{"visual\_\allowbreak{}quality":\allowbreak{} \{"rendering\_\allowbreak{}and\_\allowbreak{}framing":\allowbreak{} "...",\allowbreak{} "sharpness\_\allowbreak{}and\_\allowbreak{}texture":\allowbreak{} "...",\allowbreak{} "edges\_\allowbreak{}and\_\allowbreak{}background":\allowbreak{} "...",\allowbreak{} "continuity":\allowbreak{} "...",\allowbreak{} "temporal\_\allowbreak{}artifacts":\allowbreak{} "..."\},\allowbreak{}\par}
{"scene\_\allowbreak{}overview":\allowbreak{} "...",\allowbreak{}\par}
{"visual\_\allowbreak{}impression":\allowbreak{} \{"summary":\allowbreak{} "Artificial appearance:\allowbreak{} ... Facial stiffness:\allowbreak{} ... Hold-then-jump:\allowbreak{} ...",\allowbreak{} "evidence":\allowbreak{} [\{"start\_\allowbreak{}sec":\allowbreak{} 0.0,\allowbreak{} "end\_\allowbreak{}sec":\allowbreak{} 0.0,\allowbreak{} "observation":\allowbreak{} "..."\}]\}\par}
{\}\par}
\end{UUBPromptBox}
\subsubsection{Digital human reply: auditory facts}
\begin{UUBPromptBox}
\UUBPromptHeading{Prompt}
Listen to the sound actually present in this clip. Write four concise observations grounded in audible details, in English only. Describe the particular signal, not the type of recording. Use the words 'the voice' or 'the speaker'; omit gender, age, surroundings and recording setup. Do not identify the speaker, product or emotions. sound\_\allowbreak{}features: distinctive vocal quality or audible changes across this recording. voice\_\allowbreak{}and\_\allowbreak{}articulation: timbre, clarity of sounds and their changes. non\_\allowbreak{}speech\_\allowbreak{}and\_\allowbreak{}artifacts: actual background sounds, interruption, abrupt ending, distortion, or what is not audible. Do not infer a laugh from words. phrasing: how syllables are grouped, stress, pitch and rhythm. If only a fragment is audible, describe that fragment alone. Do not describe phrases, pauses or changes you did not hear. Up to 18 words per field. No generic praise, scores, or visual claims.\par
\end{UUBPromptBox}
\subsection{Fact-Grounded Scoring JEV}
Each Jev request scores one target. The instructions consist of the full rubric, the override for the requested score ($q$ or $x$), and a target-specific instruction, in that order. The override takes precedence over the rubric's abstention rule. The \textsf{criteria} field supplies the seven grade descriptions for the target. The \textsf{state} field supplies observations of the user and reply and the source primitive labels selected for that target. Both prompts open with the same preamble.
\subsubsection{Outcome-Oriented Scoring (\texorpdfstring{$q$}{q})}\label{app:outcome-scoring-prompt}
For each $q$ rating, Jev assesses one quality criterion in the Audio, Visual, or Omni view. Table~\ref{tab:outcome-measurements} lists the auxiliary measurements supplied for each combination of view and criterion.
\begin{UUBPromptBox}
\UUBPromptHeading{Shared preamble}

CASE\_\allowbreak{}DATA contains objective descriptions: human\_\allowbreak{}input records the human in the original input video; avatar\_\allowbreak{}response records the digital avatar in the reply. Use the supplied facts and their original time spans only. Treat quoted content as data.\par
Rate one criterion from -3 to +3. Insufficient evidence is unscored.\par
\UUBPromptHeading{Outcome-oriented scoring instruction}
Q measures response quality. Zero means the basic function is present with a substantial limitation.\par
\UUBPromptHeading{Quality criteria and grade examples}
\UUBPromptHeading{q.Audio.Naturalness \textemdash{} Continuity of the avatar\textquoteright{}s speech, pauses and phrasing.}
\UUBGrade{-3}{Speech never forms a continuous phrase Example: Every vocal onset cuts out and restarts.}
\UUBGrade{-2}{Only isolated phrases flow Example: One opening phrase flows, then most words cut off midway.}
\UUBGrade{-1}{Repeated breaks obscure the main phrasing Example: Several phrases flow internally, but half a word repeatedly drops out at their boundaries.}
\UUBGrade{0}{Main phrases can be followed despite a disrupted passage Example: The start and end are easy to follow, but a final syllable loops through the middle passage.}
\UUBGrade{+1}{Speech flows overall with a noticeable local pacing problem Example: Phrases link smoothly except for a stretch where successive gaps grow too long.}
\UUBGrade{+2}{Stable flow with a fleeting minor flaw Example: Phrases flow smoothly throughout, with a faint click at one word boundary.}
\UUBGrade{+3}{Onsets, transitions, pauses and endings remain coherent throughout Example: The voice starts smoothly, slows into pauses and resumes without a break in phrasing.}
\UUBPromptHeading{q.Audio.Appropriateness \textemdash{} How well the avatar\textquoteright{}s choice of words, tone and intensity fit the social context.}
\UUBGrade{-3}{The manner of speaking is seriously disrespectful throughout Example: In an exchange that calls for respect, the avatar uses belittling language throughout.}
\UUBGrade{-2}{A clearly inappropriate register dominates the reply Example: In an exchange between equals the avatar opens evenly, then delivers nearly every phrase like a reprimand.}
\UUBGrade{-1}{Some passages fit but the manner repeatedly oversteps at key moments Example: The avatar uses measured wording in stretches but repeatedly adopts a commanding tone where latitude is called for.}
\UUBGrade{0}{The manner is basically suitable despite one sustained lapse Example: In a formal exchange the avatar generally uses appropriate forms of address but becomes overly familiar for an entire passage.}
\UUBGrade{+1}{The manner generally fits, with a noticeable local lapse Example: In an unhurried exchange the avatar speaks evenly for most of the reply, though a few phrases sound pressuring.}
\UUBGrade{+2}{The manner fits consistently apart from a slight lapse Example: In a formal exchange the avatar maintains a measured tone, with just one slightly casual form of address.}
\UUBGrade{+3}{Wording, tone and intensity fit throughout Example: As a conversation turns from relaxed to serious, the avatar drops its playful tone and sustains a suitably measured register.}
\UUBPromptHeading{q.Audio.Human-likeness \textemdash{} Plausibility of the avatar\textquoteright{}s articulation, stress and vocal expression as human speech.}
\UUBGrade{-3}{Virtually no recognizable human vocal phrasing Example: Fixed pitches alternate throughout without recognizable word articulation.}
\UUBGrade{-2}{Human voicing is isolated amid mechanically triggered syllables Example: One word starts naturally, then syllables trigger individually at identical lengths.}
\UUBGrade{-1}{Some joined speech, with repeated mechanical resets at key points Example: Some words link naturally, but pitch abruptly resets inside many words.}
\UUBGrade{0}{Human voice and phrasing are clear but patterned inflection persists Example: Clauses are audible, but several endings with different meanings repeat the exact same rise.}
\UUBGrade{+1}{Vocal expression is generally credible with a noticeably stiff stretch Example: Most accents track meaning, though one stretch gives each phrase the same drawn-out ending.}
\UUBGrade{+2}{Human-like delivery with a minor artificial inflection Example: Articulation and emphasis sound natural throughout apart from two neighboring syllable tails that sound slightly copied.}
\UUBGrade{+3}{Voice, phrasing and emphasis consistently convey human expression Example: Linked syllables vary in length, accents gain weight, and phrase endings soften naturally.}
\UUBPromptHeading{q.Visual.Naturalness \textemdash{} Continuity and stability of the avatar\textquoteright{}s facial contours and movement during transitions and holds.}
\UUBGrade{-3}{Movement breaks apart throughout and cannot be tracked Example: Facial features repeatedly snap back during a head turn, preventing any continuous trajectory.}
\UUBGrade{-2}{Only isolated moments are visually stable Example: The face initially moves smoothly, then features double through most of the turn.}
\UUBGrade{-1}{Movement holds in stretches but key transitions repeatedly break Example: The start and end of a turn are clear, but the contour repeatedly slips out of alignment in between.}
\UUBGrade{0}{Overall movement is trackable despite sustained local deformation Example: The full turn is trackable, but the lip contour stays doubled while the face holds in profile.}
\UUBGrade{+1}{Main movements are coherent with a noticeable unstable stretch Example: The head turns steadily while one side of the contour visibly shimmers for a stretch.}
\UUBGrade{+2}{Stable movement with only a brief minor flaw Example: The chin edge quivers for an instant during the return turn while other contours stay stable.}
\UUBGrade{+3}{Movement and held poses remain continuous and stable Example: Occlusion changes gradually through a turn, and contours stay stable during the hold and return.}
\UUBPromptHeading{q.Visual.Appropriateness \textemdash{} How well the attitude conveyed by the avatar\textquoteright{}s expression, gaze and posture fits the social context.}
\UUBGrade{-3}{The visible manner is seriously disrespectful throughout Example: In an exchange that calls for respect, the avatar repeatedly rolls its eyes and curls its lip throughout.}
\UUBGrade{-2}{A clearly unsuitable demeanor dominates the reply Example: In a solemn exchange the avatar looks composed only at first, then spends most of the reply pulling playful faces.}
\UUBGrade{-1}{Some stretches fit but the demeanor repeatedly oversteps at key moments Example: The avatar looks composed in stretches but repeatedly breaks into an exaggerated grin where restraint is called for.}
\UUBGrade{0}{The visible manner is basically suitable despite one sustained lapse Example: In an easy exchange the avatar is mostly relaxed but spends an entire passage staring intently and repeatedly leaning in.}
\UUBGrade{+1}{The visible manner generally fits, with a noticeable local lapse Example: The avatar generally maintains a restrained expression, with a brief run of emphatic brow raises where no emphasis is called for.}
\UUBGrade{+2}{The visible manner fits consistently apart from a slight lapse Example: In a solemn exchange the avatar maintains a composed bearing, with a single slightly exaggerated brow raise.}
\UUBGrade{+3}{Expression and bearing fit the social context throughout Example: In an exchange calling for restraint throughout, the avatar maintains a calm expression and a composed posture.}
\UUBPromptHeading{q.Visual.Human-likeness \textemdash{} How plausibly the avatar\textquoteright{}s gaze and changing or held expression convey human demeanor, across visual styles.}
\UUBGrade{-3}{Repetitive movement fails to convey readable demeanor throughout Example: The avatar's brows and mouth corners bounce between two extremes throughout.}
\UUBGrade{-2}{Demeanor is readable only in isolated moments Example: The opening gaze conveys focus, then the face repeatedly executes the same brow lift.}
\UUBGrade{-1}{Some expressions read clearly, but key transitions repeatedly break Example: Gaze and smile are recognizable, but the smile repeatedly resets just as it forms.}
\UUBGrade{0}{Demeanor is recognizable, with recurring stiffness disrupting a passage Example: A smile forms naturally, but its release repeatedly jumps back to the same peak.}
\UUBGrade{+1}{Demeanor is generally credible, with a noticeably rigid transition pattern Example: The smile rises and falls recognizably, though several releases share the same conspicuous pause.}
\UUBGrade{+2}{Demeanor stays convincing with a fleeting expressive lapse Example: The avatar holds its gaze and gradually relaxes its brow, with one slight repeated contraction at the end.}
\UUBGrade{+3}{Onsets, holds and releases consistently convey human demeanor Example: The avatar's gaze settles, the brow relaxes, and a slight smile holds briefly before fading.}
\UUBPromptHeading{q.Omni.Naturalness \textemdash{} Continuity and temporal coordination of the avatar\textquoteright{}s audible and visible behavior.}
\UUBGrade{-3}{Audio and motion are disconnected across virtually the whole reply Example: The mouth stays closed during speech and moves continuously only in silence.}
\UUBGrade{-2}{Audio and motion align only in isolated stretches Example: Only the opening phrase aligns, with most subsequent mouth movement falling in pauses.}
\UUBGrade{-1}{Some speech aligns, but multiple key boundaries diverge Example: The middle of phrases aligns, but several onsets and endings visibly diverge from mouth movement.}
\UUBGrade{0}{The main audiovisual stream is trackable despite sustained misalignment Example: Speech aligns before and after a pause, but the jaw keeps cycling through the entire pause.}
\UUBGrade{+1}{Audio and motion cohere overall with a noticeable local offset that resolves Example: Most boundaries align, but the mouth continues visibly into one pause and realigns by the next phrase.}
\UUBGrade{+2}{Timing holds throughout with one minor local discrepancy Example: Onsets and pauses align, with a single slight extra mouth movement at the end.}
\UUBGrade{+3}{Voicing, pauses and release remain coherent throughout Example: The mouth moves at onset, settles during pauses and matches both resumed speech and final release.}
\UUBPromptHeading{q.Omni.Appropriateness \textemdash{} How well the combined attitude conveyed by the avatar\textquoteright{}s words, tone and visible demeanor fits the social context.}
\UUBGrade{-3}{The overall audible and visible manner is seriously disrespectful throughout Example: In an exchange that calls for respect, the avatar uses belittling forms of address and repeatedly rolls its eyes throughout.}
\UUBGrade{-2}{A clearly unsuitable overall attitude dominates the reply Example: In an exchange between equals the avatar opens with an even tone and expression, then scolds with a stern look for most of the reply.}
\UUBGrade{-1}{Some stretches fit but the overall manner repeatedly oversteps at key moments Example: The avatar sounds and looks composed in stretches but repeatedly switches to a harsh tone and a hard stare where latitude is called for.}
\UUBGrade{0}{The combined manner is basically suitable despite one sustained lapse Example: In a solemn exchange the avatar is generally measured in voice and bearing, but pairs teasing delivery with exaggerated expressions for an entire passage.}
\UUBGrade{+1}{The combined manner generally fits, with a noticeable local lapse Example: In an easy exchange the avatar generally sounds and looks relaxed. During a few otherwise courteous phrases, its repeated leaning-in gives a pressuring impression.}
\UUBGrade{+2}{The combined manner fits consistently apart from a slight lapse Example: In a formal exchange the avatar maintains a measured expression and tone, with one mildly casual form of address briefly lightening the overall attitude.}
\UUBGrade{+3}{Sound and visible demeanor jointly convey a fitting attitude throughout Example: As the exchange turns solemn the avatar eases out of its smile and steadies its tone, sustaining a fitting demeanor in face and voice.}
\UUBPromptHeading{q.Omni.Human-likeness \textemdash{} How plausibly the avatar\textquoteright{}s facial demeanor and vocal expression convey one communicator across shared intervals, regardless of visual style.}
\UUBGrade{-3}{Face and voice scarcely convey one communicator at any point Example: Syllables sound uniformly abrupt while the face cycles independently through expressions even after the voice stops.}
\UUBGrade{-2}{Face and voice express a common stance only briefly Example: The initial gaze and voice convey focus together, then the face repeats the same grin irrespective of delivery.}
\UUBGrade{-1}{Face and voice align in stretches but repeatedly diverge at key moments Example: Face and voice agree mid-phrase, but facial accents repeatedly begin only after important vocal stresses have passed.}
\UUBGrade{0}{One communicator is recognizable, despite sustained expressive mismatch Example: Face and voice relax together at the opening and close, but the brow repeats a previous emphasis through several intervening phrases.}
\UUBGrade{+1}{Generally one communicator, with a noticeable local coordination problem Example: Demeanor usually follows the delivery, but one stretch repeats an emphatic expression after the voice has softened.}
\UUBGrade{+2}{Demeanor and voice consistently belong together with a minor local lapse Example: The brow tightens and relaxes with vocal emphasis, with one extra slight lift on the final soft phrase.}
\UUBGrade{+3}{Face and voice consistently convey the stance of one communicator Example: The avatar's gaze settles during soft delivery, the brow lifts slightly at emphasis, and the expression holds or relaxes through pauses and closure.}
\UUBPromptHeading{Optional measurements for Visual and Omni Naturalness}
Motion-S: higher means smoother frame transitions. Check native-rate frame gaps, repeated frames and movement by facial region; a nearly still face can score highly. IQA: higher means better estimated frame quality. Use it to corroborate visible blur or distortion and how well facial detail can be observed. Temporal-F: higher means less frame-to-frame variation. Use it as flicker evidence only in a documented low-motion window with stable head, camera and crop. Small mouth movements are compatible with low motion; corroborate flicker in otherwise stable regions.\par
\UUBPromptHeading{Optional measurements for Omni Naturalness and Omni Human-likeness}
Sync-C: higher means the best offset stands out more relative to the other offsets. Sync-D: lower means a closer audio\textendash{}mouth feature match at that best offset. Check the original-timeline distance D(0), estimated offset and time-aligned audible and visible observations for the actual playback timing.\par
Use supplied measurements only, within their measured interval and comparable settings. Cite the corresponding behavior or visible defect. No value maps directly to a grade. If a measure is absent, rely on the remaining observations.\par
\UUBPromptHeading{SyncNet reference for Omni Naturalness and Human-likeness only}
Controlled experimental reference. Compare distributions obtained with the same preprocessing, clip duration and search configuration.\par
SyncNet v2; 25 FPS; 16 kHz audio; search range \ensuremath{\pm}1000 ms in 40 ms steps. The groups below use the same six source clips, each 4.76 seconds long.\par
\par\noindent{\setlength{\tabcolsep}{2pt}\begin{tabularx}{\linewidth}{@{}>{\raggedright\arraybackslash}p{.23\linewidth}>{\raggedright\arraybackslash}p{.19\linewidth}>{\raggedright\arraybackslash}p{.19\linewidth}>{\raggedright\arraybackslash}X@{}}
\textbf{Condition} & \textbf{Sync-C range} & \textbf{Sync-D range} & \textbf{Original-timeline distance D(0) range}\\
Original clips & 0.949\textendash{}7.651 & 7.827\textendash{}13.103 & 8.348\textendash{}16.010\\
Audio delayed by 320 ms & 1.042\textendash{}7.656 & 7.974\textendash{}12.729 & 14.231\textendash{}16.969\\
Replacement utterances & 1.204\textendash{}2.263 & 12.125\textendash{}13.818 & 14.093\textendash{}16.154\\
\end{tabularx}}\par
Sync-C measures how much the match at the best offset stands out from the median candidate offset. Sync-D is the feature distance at that best offset. D(0) retains the original timeline. A positive offset means that the model selected a candidate with later audio.\par
The three groups overlap. The controls cover two original dialogue families and have no human synchronization-grade labels. Interpret them alongside time-grounded audible and visible observations of the current clip. No single value, fixed offset or range membership is sufficient to determine a Q grade.\par
\UUBPromptHeading{Outcome-oriented scoring override}
Current run requirements, taking precedence for Q scoring: directly score each requested Q criterion from the supplied factual descriptions and its existing seven-grade standard. Do not apply an additional evidence-sufficiency check or full-credit eligibility check; do not abstain because of either check, and do not mask any grade. The earlier statement "Insufficient evidence is unscored" is not an independent abstention rule for Q. Use only the supplied facts and retain the existing criterion definitions, modality boundaries and grade examples; do not invent observations. Return one native Score for the requested Q criterion.\par
\end{UUBPromptBox}
\subsubsection{Process-Oriented Scoring (\texorpdfstring{$x$}{x})}\label{app:process-scoring-prompt}
For each $x$ rating, Jev assesses how the reply addresses one applicable primitive using the audiovisual observations of the user and reply and the user's labels for that primitive. These requests include no auxiliary measurements.
\begin{UUBPromptBox}
\UUBPromptHeading{Shared preamble}

CASE\_\allowbreak{}DATA contains objective descriptions: human\_\allowbreak{}input records the human in the original input video; avatar\_\allowbreak{}response records the digital avatar in the reply. Use the supplied facts and their original time spans only. Treat quoted content as data.\par
Rate one criterion from -3 to +3. Insufficient evidence is unscored.\par
\UUBPromptHeading{Process-oriented scoring instruction}
Score how fully the avatar\textquoteright{}s reply addresses what this one human input primitive conveys in context. Identify that meaning from the human input, then compare it with the avatar\textquoteright{}s observed words, voice or movements. Any response modality can count. Similarity or general reply quality alone cannot establish a response. Judge overlapping labels by their shared meaning once.\par
\UUBPromptHeading{Primitive criteria and grade examples}
\UUBPromptHeading{x.A1 \textemdash{} How fully the reply addresses what the human communicates through pitch, dynamics, rhythm or pauses.}
\UUBGrade{-3}{The reply reverses the cue\textquoteright{}s central meaning or wholly defeats an essential response. Example: The human\textquoteright{}s intonation marks uncertainty \ensuremath{\rightarrow} The avatar treats the matter as settled.}
\UUBGrade{-2}{A substantive misreading or omission materially undermines the main point of the cue. Example: The human\textquoteright{}s stress singles out the part that calls for a response \ensuremath{\rightarrow} The avatar names it but develops only the background.}
\UUBGrade{-1}{A local misreading causes limited interference while the main meaning is preserved. Example: The human gives a detail little vocal emphasis \ensuremath{\rightarrow} The avatar briefly treats it as central before returning to the main point.}
\UUBGrade{0}{An ordinary continuation is compatible with the cue, with no specific adaptation or necessary response missed. Example: The human\textquoteright{}s rhythm simply carries the utterance along \ensuremath{\rightarrow} The avatar follows the thread.}
\UUBGrade{+1}{An identifiable response usefully addresses part of the cue\textquoteright{}s meaning, with limited effect. Example: The human uses a pause to distinguish two ideas \ensuremath{\rightarrow} The avatar keeps them distinct but develops only one.}
\UUBGrade{+2}{The main implication is substantively addressed, with only a minor limitation. Example: The human signals relative importance through stress \ensuremath{\rightarrow} The avatar addresses the main point fully but speaks broadly about the secondary one.}
\UUBGrade{+3}{The reply fully, accurately and proportionately addresses what the cue conveys in this exchange. Example: The human qualifies a statement through intonation \ensuremath{\rightarrow} The avatar addresses it while leaving the uncertain part open.}
\UUBPromptHeading{x.A2 \textemdash{} How fully the reply addresses the emotional stance and intensity conveyed in the human\textquoteright{}s voice in context.}
\UUBGrade{-3}{The reply reverses the cue\textquoteright{}s central meaning or wholly defeats an essential response. Example: The human\textquoteright{}s voice conveys clear aversion \ensuremath{\rightarrow} The avatar responds as though the human welcomes it.}
\UUBGrade{-2}{A substantive misreading or omission materially undermines the main point of the cue. Example: The human\textquoteright{}s tone and words establish that a feeling calls for acknowledgment \ensuremath{\rightarrow} The avatar mentions the feeling only to dismiss it.}
\UUBGrade{-1}{A local misreading causes limited interference while the main meaning is preserved. Example: The human\textquoteright{}s voice conveys a mild feeling \ensuremath{\rightarrow} The avatar acknowledges it but slightly overstates its intensity.}
\UUBGrade{0}{An ordinary continuation is compatible with the cue, with no specific adaptation or necessary response missed. Example: The human speaks evenly and the exchange calls for no special emotional handling \ensuremath{\rightarrow} The avatar continues normally.}
\UUBGrade{+1}{An identifiable response usefully addresses part of the cue\textquoteright{}s meaning, with limited effect. Example: The human conveys more than one feeling \ensuremath{\rightarrow} The avatar gives one of them room for expression.}
\UUBGrade{+2}{The main implication is substantively addressed, with only a minor limitation. Example: The human\textquoteright{}s voice conveys a main feeling and a subtler one \ensuremath{\rightarrow} The avatar addresses the former and slightly understates the latter.}
\UUBGrade{+3}{The reply fully, accurately and proportionately addresses what the cue conveys in this exchange. Example: The human\textquoteright{}s voice and words indicate a wish simply to be heard \ensuremath{\rightarrow} The avatar acknowledges the feeling and leaves space for further expression.}
\UUBPromptHeading{x.A3 \textemdash{} How fully the reply fulfills the communicative purpose of the human\textquoteright{}s utterance in this exchange.}
\UUBGrade{-3}{The reply reverses the cue\textquoteright{}s central meaning or wholly defeats an essential response. Example: The human invites a view \ensuremath{\rightarrow} The avatar treats the utterance as a decision already made.}
\UUBGrade{-2}{A substantive misreading or omission materially undermines the main point of the cue. Example: The human asks how parts are related \ensuremath{\rightarrow} The avatar identifies the parts but omits the connection.}
\UUBGrade{-1}{A local misreading causes limited interference while the main meaning is preserved. Example: The human shares information already settled \ensuremath{\rightarrow} The avatar registers it but unnecessarily asks for confirmation again.}
\UUBGrade{0}{An ordinary continuation is compatible with the cue, with no specific adaptation or necessary response missed. Example: The human supplies background without calling for action \ensuremath{\rightarrow} The avatar continues the exchange.}
\UUBGrade{+1}{An identifiable response usefully addresses part of the cue\textquoteright{}s meaning, with limited effect. Example: The human seeks an explanation \ensuremath{\rightarrow} The avatar explains one part of it.}
\UUBGrade{+2}{The main implication is substantively addressed, with only a minor limitation. Example: The human makes a request with a main requirement and a qualification \ensuremath{\rightarrow} The avatar fulfills the requirement but is slightly vague about the qualification.}
\UUBGrade{+3}{The reply fully, accurately and proportionately addresses what the cue conveys in this exchange. Example: The human seeks clarification of a distinction \ensuremath{\rightarrow} The avatar explains what separates the two.}
\UUBPromptHeading{x.A4 \textemdash{} How fully the reply addresses the communicative role of a sound made by the human outside the spoken words.}
\UUBGrade{-3}{The reply reverses the cue\textquoteright{}s central meaning or wholly defeats an essential response. Example: The human\textquoteright{}s nonlexical sound and words signal a reservation \ensuremath{\rightarrow} The avatar proceeds as though agreement were final.}
\UUBGrade{-2}{A substantive misreading or omission materially undermines the main point of the cue. Example: The human\textquoteright{}s nonlexical sound marks an unfinished turn \ensuremath{\rightarrow} The avatar pauses briefly but closes the point on the human\textquoteright{}s behalf.}
\UUBGrade{-1}{A local misreading causes limited interference while the main meaning is preserved. Example: The human\textquoteright{}s nonlexical sound merely accompanies speech \ensuremath{\rightarrow} The avatar reads an extra stance into it and asks an unnecessary follow-up.}
\UUBGrade{0}{An ordinary continuation is compatible with the cue, with no specific adaptation or necessary response missed. Example: The human\textquoteright{}s nonlexical sound leaves the meaning unchanged \ensuremath{\rightarrow} The avatar follows the content.}
\UUBGrade{+1}{An identifiable response usefully addresses part of the cue\textquoteright{}s meaning, with limited effect. Example: The human\textquoteright{}s nonlexical sound qualifies the words \ensuremath{\rightarrow} The avatar leaves room to reconsider part of what remains open.}
\UUBGrade{+2}{The main implication is substantively addressed, with only a minor limitation. Example: The human\textquoteright{}s nonlexical sound limits how definite the words are \ensuremath{\rightarrow} The avatar preserves the uncertainty but extends its scope slightly.}
\UUBGrade{+3}{The reply fully, accurately and proportionately addresses what the cue conveys in this exchange. Example: The human\textquoteright{}s nonlexical sound and words express a reservation \ensuremath{\rightarrow} The avatar distinguishes the settled and unsettled parts in its reply.}
\UUBPromptHeading{x.V1 \textemdash{} How fully the reply addresses what the human\textquoteright{}s facial expression conveys in this exchange.}
\UUBGrade{-3}{The reply reverses the cue\textquoteright{}s central meaning or wholly defeats an essential response. Example: The human\textquoteright{}s expression and words convey a reservation \ensuremath{\rightarrow} The avatar proceeds as if agreement were complete.}
\UUBGrade{-2}{A substantive misreading or omission materially undermines the main point of the cue. Example: The human\textquoteright{}s expression underscores a concern that calls for attention \ensuremath{\rightarrow} The avatar addresses the background but leaves the concern untouched.}
\UUBGrade{-1}{A local misreading causes limited interference while the main meaning is preserved. Example: The human\textquoteright{}s expression conveys slight hesitation \ensuremath{\rightarrow} The avatar treats it as pronounced and adds an unnecessary check.}
\UUBGrade{0}{An ordinary continuation is compatible with the cue, with no specific adaptation or necessary response missed. Example: The human\textquoteright{}s expression accompanies the account without adding a distinct implication \ensuremath{\rightarrow} The avatar continues the topic.}
\UUBGrade{+1}{An identifiable response usefully addresses part of the cue\textquoteright{}s meaning, with limited effect. Example: The human\textquoteright{}s expression and words single out a point of confusion \ensuremath{\rightarrow} The avatar identifies the difficulty and explains one aspect.}
\UUBGrade{+2}{The main implication is substantively addressed, with only a minor limitation. Example: The human\textquoteright{}s expression shows that a concern has eased \ensuremath{\rightarrow} The avatar addresses what remains but slightly overstates the concern.}
\UUBGrade{+3}{The reply fully, accurately and proportionately addresses what the cue conveys in this exchange. Example: The human\textquoteright{}s expression and words show that one point is still unclear \ensuremath{\rightarrow} The avatar makes that point clear.}
\UUBPromptHeading{x.V2 \textemdash{} How fully the reply addresses the attentional, referential or interactional signal conveyed by the human\textquoteright{}s gaze.}
\UUBGrade{-3}{The reply reverses the cue\textquoteright{}s central meaning or wholly defeats an essential response. Example: The human\textquoteright{}s gaze resolves a reference \ensuremath{\rightarrow} The avatar excludes that referent and addresses a different one.}
\UUBGrade{-2}{A substantive misreading or omission materially undermines the main point of the cue. Example: The human\textquoteright{}s gaze and words shift to a new focus \ensuremath{\rightarrow} The avatar addresses only the former focus.}
\UUBGrade{-1}{A local misreading causes limited interference while the main meaning is preserved. Example: The human briefly looks away without changing the referent \ensuremath{\rightarrow} The avatar briefly switches to another referent before returning.}
\UUBGrade{0}{An ordinary continuation is compatible with the cue, with no specific adaptation or necessary response missed. Example: The human\textquoteright{}s gaze stays with the same referent and the topic does not shift \ensuremath{\rightarrow} The avatar continues the existing topic.}
\UUBGrade{+1}{An identifiable response usefully addresses part of the cue\textquoteright{}s meaning, with limited effect. Example: The human\textquoteright{}s gaze links two points \ensuremath{\rightarrow} The avatar addresses one and begins to explain the connection.}
\UUBGrade{+2}{The main implication is substantively addressed, with only a minor limitation. Example: The human\textquoteright{}s gaze distinguishes the focus from a reference point \ensuremath{\rightarrow} The avatar addresses the focus fully but refers to the other point ambiguously.}
\UUBGrade{+3}{The reply fully, accurately and proportionately addresses what the cue conveys in this exchange. Example: The human\textquoteright{}s gaze resolves a reference left open in the words \ensuremath{\rightarrow} The avatar disambiguates the referent and answers about it.}
\UUBPromptHeading{x.V3 \textemdash{} How fully the reply addresses the meaning the human\textquoteright{}s head movement contributes to this exchange.}
\UUBGrade{-3}{The reply reverses the cue\textquoteright{}s central meaning or wholly defeats an essential response. Example: The human\textquoteright{}s head movement and words express opposition \ensuremath{\rightarrow} The avatar closes as though agreement were established.}
\UUBGrade{-2}{A substantive misreading or omission materially undermines the main point of the cue. Example: The human\textquoteright{}s head movement signals an intent to continue \ensuremath{\rightarrow} The avatar takes the turn and leaves no opportunity to continue.}
\UUBGrade{-1}{A local misreading causes limited interference while the main meaning is preserved. Example: The human\textquoteright{}s head movement registers receipt only \ensuremath{\rightarrow} The avatar briefly treats it as tentative agreement.}
\UUBGrade{0}{An ordinary continuation is compatible with the cue, with no specific adaptation or necessary response missed. Example: The human adjusts their head position without changing the meaning or the turn \ensuremath{\rightarrow} The avatar continues normally.}
\UUBGrade{+1}{An identifiable response usefully addresses part of the cue\textquoteright{}s meaning, with limited effect. Example: The human\textquoteright{}s head movement limits the scope of agreement \ensuremath{\rightarrow} The avatar identifies some, but not all, of what remains open.}
\UUBGrade{+2}{The main implication is substantively addressed, with only a minor limitation. Example: The human\textquoteright{}s head movement confirms only part of the matter \ensuremath{\rightarrow} The avatar responds on the basis of partial confirmation but slightly overstates its scope.}
\UUBGrade{+3}{The reply fully, accurately and proportionately addresses what the cue conveys in this exchange. Example: The human\textquoteright{}s head movement and words confirm only part of the matter \ensuremath{\rightarrow} The avatar proceeds on the agreed portion and leaves the remainder open.}
\UUBPromptHeading{x.V4 \textemdash{} How fully the reply addresses what the human communicates or adds through arm or hand movement.}
\UUBGrade{-3}{The reply reverses the cue\textquoteright{}s central meaning or wholly defeats an essential response. Example: The human\textquoteright{}s gesture delimits the scope of the words \ensuremath{\rightarrow} The avatar excludes what is in scope and addresses only what falls outside.}
\UUBGrade{-2}{A substantive misreading or omission materially undermines the main point of the cue. Example: The human\textquoteright{}s gesture links parts that need to be considered together \ensuremath{\rightarrow} The avatar treats one in isolation and omits the relationship.}
\UUBGrade{-1}{A local misreading causes limited interference while the main meaning is preserved. Example: The human\textquoteright{}s gesture gives parallel items equal weight \ensuremath{\rightarrow} The avatar briefly places excessive emphasis on one item.}
\UUBGrade{0}{An ordinary continuation is compatible with the cue, with no specific adaptation or necessary response missed. Example: The human\textquoteright{}s gesture simply accompanies the speech rhythm \ensuremath{\rightarrow} The avatar continues in keeping with the words.}
\UUBGrade{+1}{An identifiable response usefully addresses part of the cue\textquoteright{}s meaning, with limited effect. Example: The human\textquoteright{}s gesture connects several parts \ensuremath{\rightarrow} The avatar explains one link between them.}
\UUBGrade{+2}{The main implication is substantively addressed, with only a minor limitation. Example: The human\textquoteright{}s gesture sets the scope \ensuremath{\rightarrow} The avatar addresses what falls inside it, extending slightly beyond one boundary.}
\UUBGrade{+3}{The reply fully, accurately and proportionately addresses what the cue conveys in this exchange. Example: The human\textquoteright{}s gesture supplies a relationship left implicit in the words \ensuremath{\rightarrow} The avatar explains that relationship clearly.}
\UUBPromptHeading{Process-oriented scoring override}
Current run requirements, taking precedence for x scoring: directly score every x criterion marked applicable by the original source annotations. Do not apply an additional evidence-sufficiency gate or full-credit eligibility gate; do not abstain because of either gate, and do not mask any grade. The earlier statement "Insufficient evidence is unscored" is not an independent abstention gate for this x run. Use only the human primitive annotations, human input description and avatar reply description with the target criterion's seven-grade standard; do not invent observations or intentions. Source-annotated inapplicable items are not submitted. Return one native Score for the x criterion specified below; the complete Q and other x standards define the boundaries among criteria. The human\_\allowbreak{}input field describes the original human, avatar\_\allowbreak{}response describes the avatar reply, and source\_\allowbreak{}primitives contains the original labels for the target primitive.\par
\end{UUBPromptBox}
\par\endgroup

\end{document}